\documentclass[10pt,twocolumn,letterpaper]{article}

\usepackage{iccv}
\usepackage{times}
\usepackage{epsfig}
\usepackage{graphicx}
\usepackage{amsmath}
\usepackage{amssymb}
\usepackage{booktabs} 
\usepackage{subcaption}
\usepackage{stackengine}
\usepackage{resizegather}
\usepackage{booktabs} 
\usepackage{xcolor}

\usepackage{titling}
\predate{} 
\postdate{}
\date{}

\usepackage[breaklinks=true,bookmarks=false]{hyperref}

\iccvfinalcopy 

\def\iccvPaperID{3768} 

\ificcvfinal\fi

\begin{document}

\title{ Bridging the Domain Gap for Ground-to-Aerial Image Matching
}

\author{Krishna Regmi
and
Mubarak Shah\\
Center for Research in Computer Vision, University of Central Florida\\
{\tt\small krishna.regmi7@gmail.com, shah@crcv.ucf.edu}
}

\maketitle
\ificcvfinal\thispagestyle{empty}\fi

\begin{abstract}
The visual entities in cross-view (e.g. ground and aerial) images exhibit drastic domain changes due to the differences in viewpoints each set of images is captured from. Existing state-of-the-art methods address the problem by learning view-invariant images descriptors.
We propose a novel method for solving this task by exploiting the generative powers of conditional GANs to synthesize an aerial representation of a ground-level panorama query and use it to minimize the domain gap between the two views. The synthesized image being from the same view as the reference (target) image, helps the network to preserve important cues in aerial images following our Joint Feature Learning approach. 
We fuse the complementary features from a synthesized aerial image with the original  ground-level panorama features to obtain a robust query representation. 
In addition, we employ multi-scale feature aggregation in order to preserve image representations at different  scales useful for solving this complex task. Experimental results show that our proposed approach performs significantly better than the state-of-the-art methods on the challenging CVUSA dataset in terms of top-1 and top-1\% retrieval accuracies. 
Furthermore, we evaluate the generalization of the proposed method for urban landscapes on  our newly collected   cross-view localization dataset with geo-reference information.

\end{abstract}


\section{Introduction} 

Estimating the geo-location of an image has been tackled as an image-matching task, where the query image is compared against a database of reference images with known locations. Traditionally, the matching has been conducted between images taken from the same view, primarily street-view \cite{Hays:2008:im2gps, Sattler_2016_CVPR, DBLP:journals/pami/ZamirS14}, which have a high degree of visual similarity in terms of  scene contents.
Since these ground level reference images are typically concentrated around urban areas with more human accessibility, the applicability of the method is limited to those regions. With the availability of aerial images from Google maps, Bing maps, etc. that cover the earth surface densely, researchers have lately explored the prospect of cross-view image matching \cite{Hu_2018_CVPR, Lin_2013_CVPR, workman2015wide}, where the query ground image is matched against aerial images. This comes with additional challenges due to variation in viewpoints between the ground and aerial images, which capture the same scene  differently in two views. This motivates us to explore transforming the query street-view image into aerial view, so that the transformed image has scene representations similar to the images it is matched against.

\begin{figure}[]
\begin{center}
   \includegraphics[width= 0.99\linewidth]{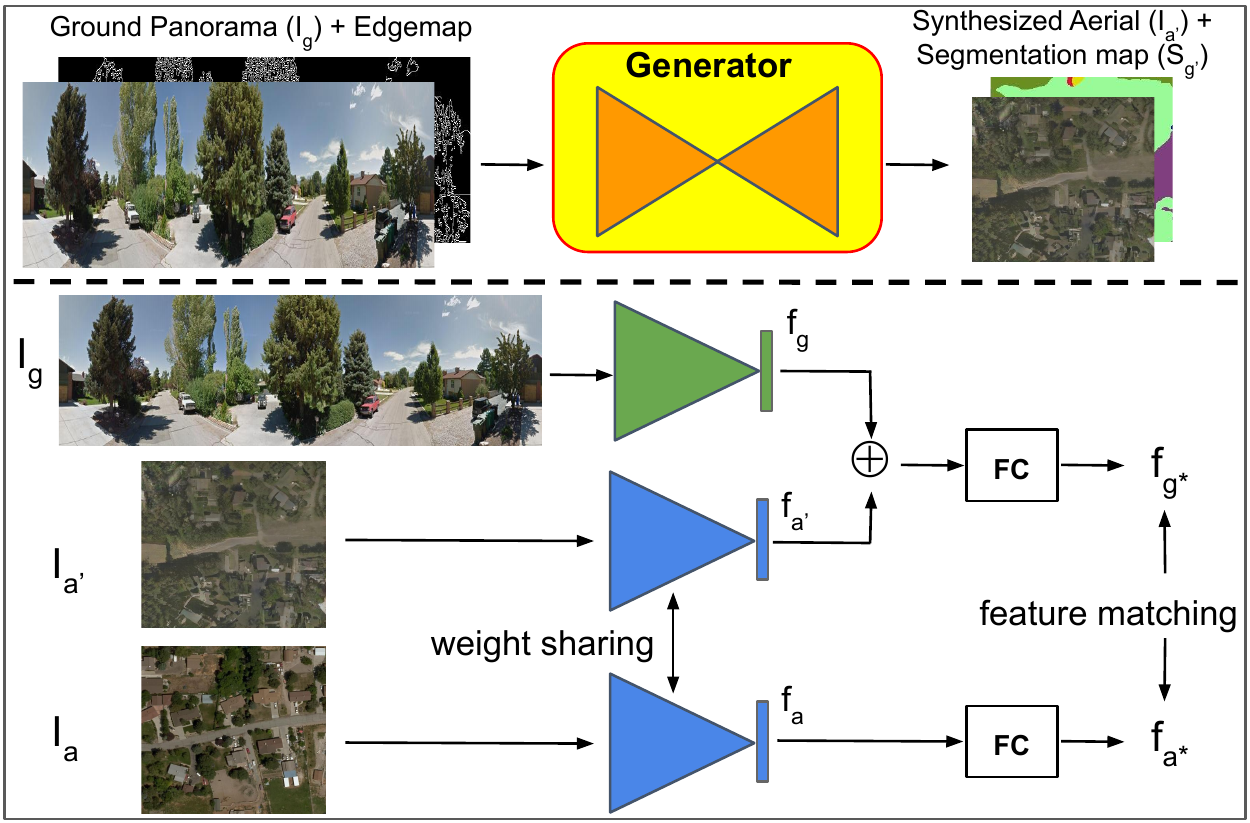}
  \vspace{-5pt}
\end{center}
   \caption{\small 
   The ground panorama query and its edgemap are inputs to the Generator (X-Fork \cite{Regmi_2018_CVPR}) network to synthesize aerial image $I_{a'}$  and its segmentation map (shown in upper panel). We then jointly learn the feature representations for image triads ($I_g$, $I_{a'}$ and $I_a$). The features $f_g$ for $I_g$ and $f_{a'}$ for $I_{a'}$ are fused, followed by fully-connected operation to obtain a robust query representation $f_{g^*}$ and is matched with the aerial feature representation $f_{a^*}$ (shown in lower panel).  
}
\label{fig:intro_new}
  \vspace{-10pt}
\end{figure}

The recent success of Generative Adversarial Networks (GANs) \cite{goodfellow2014generative} in synthesizing realistic images from randomly sampled noise vectors \cite{radford2015unsupervised} or conditional variables such as text \cite{pmlr-v48-reed16, han2017stackgan}, images \cite{pix2pix2017, Regmi_2018_CVPR}, labels \cite{DBLP:journals/corr/MirzaO14}, etc. has inspired us to frame the problem as viewpoint translation followed by feature matching. Moreover, GANs have been used for domain transfer problems as in \cite{pmlr-v70-kim17a, CycleGAN2017} to learn the mapping between different domain representations. Recent cross-view synthesis works by \cite{Regmi_2018_CVPR, REGMI2019, Deng:2018:LDG:3274895.3274969, zhu2018generative} are successful in transforming the images between aerial and street views. In this paper, we address the following problem:  given a  ground-level panorama retrieve matching aerial images.
In order to solve this problem, we take a next step to synthesize aerial images from {\em ground-level panorama} and  use them for image retrieval.

The complexity of the cross-view image synthesis problem and its challenges are well-known. Thus, the synthesized images cannot be relied on to completely replace the query ground-level image to solve the matching task. Therefore, we propose a framework as shown in Figure \ref{fig:intro_new} to incorporate the synthesized image into the matching pipeline as auxiliary information in order to bridge the existing domain gap between aerial and ground views. We attempt to learn representations for aerial reference images that are similar to their corresponding ground level images, as well as the synthesized aerial images. Since the synthesized aerial images are transformed representations of street-view (ground) images, we expect them to contain representative features.
By learning representations in this manner, the synthesized aerial images force the network to minimize the distance between feature representations of aerial images and street-view images.
Additionally, we hypothesize that some features of aerial images are better learned by considering synthesized aerial images rather than street-view images. Thus, the joint training of these 
image triads (ground, synthesize aerial from ground, and corresponding real aerial) will help the aerial stream retain important cues that would have otherwise been lost in cross-view training. We fuse the learned complementary feature representations of synthesized images with query image features to obtain a robust representation that we use for our image matching task.

The features extracted at different layers of deep neural networks capture varying levels of semantic information 
of the input image. For the image matching task, which is considerably more challenging than a standard classification problem, 
we exploit the inherent multi-scale pyramidal structure of features at multiple layers of deep neural networks and aggregate them to obtain a better image representation.

In summary, this paper makes the following contributions. We propose a novel approach to leverage aerial images synthesized using GANs to extract complementary features for cross-view image matching. We incorporate the edgemaps, in addition to semantic segmentation which is typically used, together with the input images to improve the cross-view synthesis by providing cues on object shapes and boundaries to the network. The synthesized images bridge the domain gap between cross-view images. The joint training of image triads using auxiliary loss helps improve the network training. 
The proposed feature fusion strategy demonstrates the capabilities of GANs for constructive training and complementary feature learning. 
Lastly, we show that aggregating features from multiple convolutional layers at different resolutions greatly helps preserve coarse to fine latent representations necessary for complex cross-view matching task. 
Our extensive experiments show that the proposed joint feature learning method outperforms the state-of-the-art methods on CVUSA dataset \cite{zhai2017crossview} and with feature fusion, we obtain significant improvements on top-1 and top-10 retrieval accuracies. 

\section{Related Works} 

\subsection{Domain Transfer and GANs}

GANs are very popular in domain transfer tasks. In the works reported in \cite{pix2pix2017, CycleGAN2017, pmlr-v70-kim17a, yi2017dualgan, elfeki2018third}, image mapping between two domains; source and target domains is learnt. Augmented CycleGAN \cite{almahairi2018augmented}, StarGAN \cite{Choi_2018_CVPR} have explored many-to-many cross-domain mappings.   

Cross-view relations have been explored in \cite{zhai2017crossview, Regmi_2018_CVPR,ghouaiel2016coupling} with more challenging settings of aerial and ground views, where there is minimum semantic and viewpoint overlap between the objects in the images. Cross-view image synthesis between these contrasting domains has attracted wide interests lately \cite{Regmi_2018_CVPR,REGMI2019,Deng:2018:LDG:3274895.3274969, zhu2018generative} with the popularity of GANs; these works 
have been successful in image translation between aerial and ground-level cropped (single camera) images. Zhai \textit{et al.} \cite{zhai2017crossview} explored the possibility of synthesizing ground-level panorama from ground semantic layouts wherein the layouts were predicted from the semantic maps of the aerial images. Here, we directly transform the ground level panorama to aerial view and use them for cross-view image matching task.


\subsection{Multi-scale Feature Aggregation} Features at different layers of deep neural networks are essentially the multi-resolution features of the same image. Abundance of literature has explored features at multiple scales \cite{honari2016recombinator, newell2016stacked, DBLP:journals/corr/RonnebergerFB15, long2015fully, lin2017refinenet} for applications like key-point detection, human pose estimation, semantic segmentation. 
FPN \cite{lin2017feature}, HyperNet \cite{kong2016hypernet}, ION \cite{bell2016inside} explored multi-scale features  for object detection. Earlier, Hypercolumns \cite{hariharan2015hypercolumns} were created from multi-layer features and used for object segmentation and localization. Building upon this work, we also aggregate the features at multiple scales to efficiently obtain robust representation of the images.


\subsection{Image Geolocalization} 
Image geolocalization has been tackled as an image matching task \cite{arandjelovic2016netvlad, Hays2015LargeScaleIG, 10.1007/978-3-642-15561-1_19} in computer vision community. Early works in geolocalization \cite{DBLP:journals/pami/ZamirS14, torii2011visual, Sattler_2016_CVPR, zemene2019large} matched images in the same view; a query street-view image is compared against the reference street-view images using hand-crafted features. Hays \etal \cite{Hays:2008:im2gps} proposed a data-driven approach to estimate the distribution over geographical location from a single image. 

Cross-view matching has been explored by several recent works \cite{Lin_2013_CVPR, shan2014accurate, Hu_2018_CVPR, workman2015location, Vo2016} using both hand-crafted features as well as deep networks. Bansal \etal \cite{10.1007/978-3-642-33863-2_18} explored facade matching. Tian \etal \cite{tian2017cross} matched building features in oblique views. Recent work by \cite{Hu_2018_CVPR} exploit the NetVLAD \cite{arandjelovic2016netvlad} to obtain view-invariant descriptors for cross-view pairs and use them for matching.


In this work, we exploit the synthesized aerial images as complementary source of information for better scene understanding of street-view images to solve cross-view matching task, rather than just learning view-invariant features as in the previous approaches. 

\section{Method}
We propose a novel method to bridge the domain gap between street-view and aerial images by leveraging the synthesized aerial images using GANs.
We learn the representations of synthesized aerial images jointly with ground and aerial image representations. 
Additionally, we fuse the complementary representations of ground images with the representations of their corresponding synthesized aerial images to learn robust query representations of ground images. Also, we exploit the edgemaps of input images to provide GANs with the notion of object shapes and boundaries and facilitate the cross-view image synthesis. 

The organization of the rest of this section is as follows. In the next subsection, we briefly describe how GANs are used for cross-view image synthesis, followed by joint feature learning, and finally feature fusion is described.

\subsection{\label{gan}Cross-View Image Synthesis} 

We adopt X-Fork generator architecture of \cite{Regmi_2018_CVPR} to train the GAN for cross-view image synthesis. The X-Fork is a multi-task learning architecture that synthesizes cross-view image as well as semantic segmentation map. 
We make the following modifications to the X-Fork architecture. Since our input is panorama (rectangular in shape), the feature maps at the bottleneck are also rectangular (1 $\times$ 4). 
We reshape the features into squares (2 $\times$ 2), and then apply multiple upconvolution operations to generate 512 $\times$ 512 resolution aerial images. 
Next, we exploit the edgemaps of input images that outline the objects present in the images. We employ Canny Edge Detection \cite{canny1987computational} to obtain the edgemaps of the inputs. The edgemap is stacked together with the panorama, along the channels to create a 4-channel input; 3 channels for RGB image and 1 channel for edgemap. The output is an RGB image and its segmentation map in aerial view. We utilize the synthesized aerial images in joint feature learning experiments.


\begin{figure}
\begin{tabular}{c}
\subcaptionbox{Joint Feature Learning: Inputs to this network are $I_g$ and $I_a$ and outputs are $f_{g}$ and $f_{a}$. 
Employing auxiliary loss between $f_{a'}$ and $f_a$ helps to pull features $f_g$ and $f_a$ closer, and minimize the domain gap between two features than when training  two-stream network on $(I_g, I_a)$ pairs. The branch in the middle (dotted box filled with cyan) is used during the training only.  
\label{fig:aux-loss}} {\includegraphics[width=1.0\linewidth]{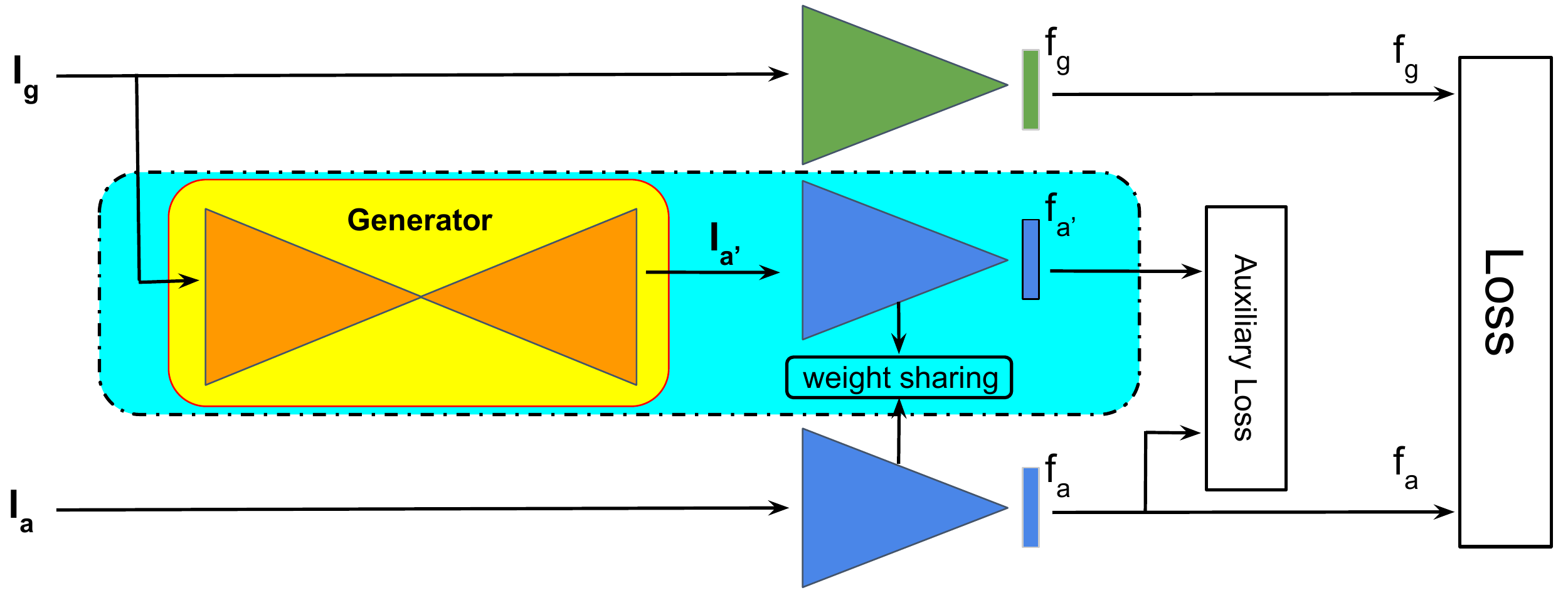}}\\ \\
\subcaptionbox{Feature Fusion. Inputs to this network are $I_g$ and $I_a$ and outputs are $f_{g^*}$ and $f_{a^*}$. $f_g$ and $f_{a'}$ are  concatenated and passed through fully-connected layer (FC) to get their fused representation $f_g*$. 
Similarly, $f_a$ is mapped to $f_{a^*}$, a representation closer to $f_{g^*}$. 
\label{fig:fusion-net}} {\includegraphics[width=1.0\linewidth]{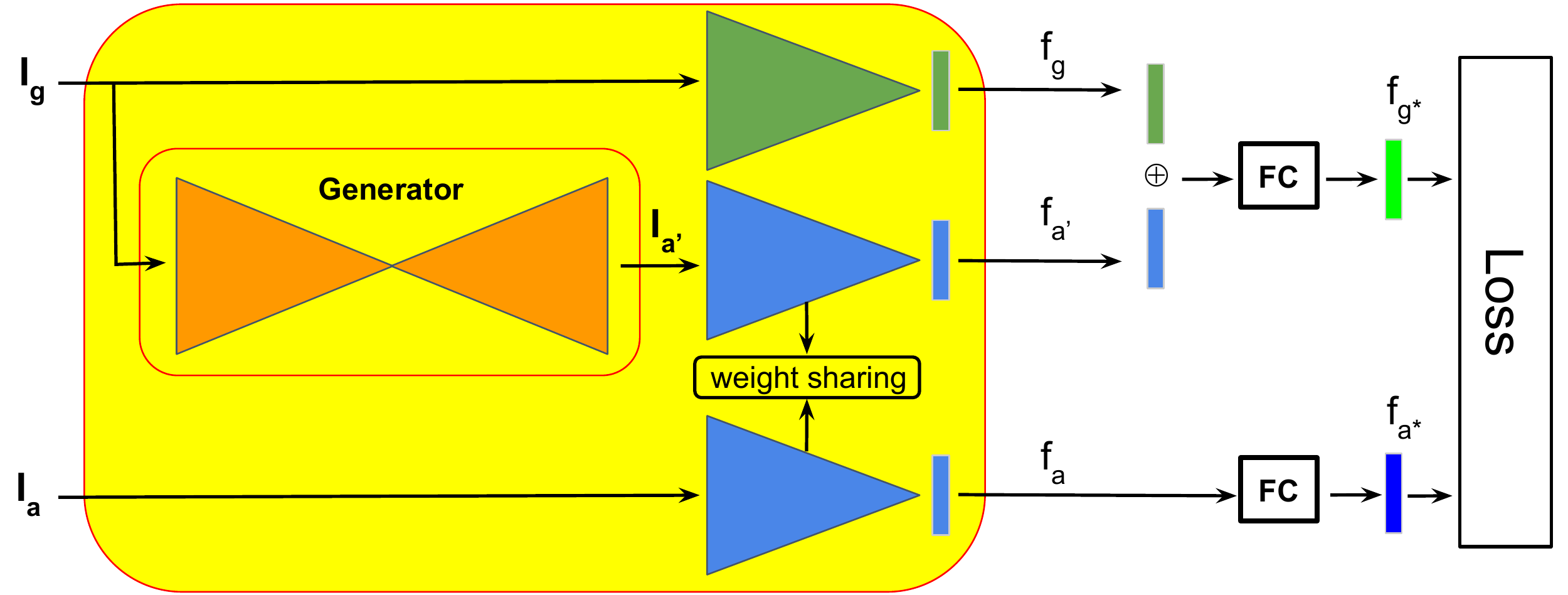}}

\end{tabular}
\caption{\small Architectures for our proposed approaches. The green and blue triangles are encoders for ground and aerial images respectively, with the network layer definition explained in subsection \ref{impl_details}. The parameters for networks shown in shade of yellow are frozen during the training. 
}
\label{fig:nw_arch}
 \vspace{-10pt}

\end{figure}


\subsection{Joint Feature Learning} 
We propose to learn the representations for image triads: query ground panorama, $I_g$, synthesized aerial image,  $I_{a'}$ from ground panorama and aerial image  $I_a$  jointly, so that the synthesized aerial image representations $f_{a'}$ pushes the image representations $f_g$ and $f_a$ closer to each other. 

The joint feature learning architecture is shown in Figure \ref{fig:aux-loss}. The encoder blocks are shown in green (for ground image) and blue (for aerial images) triangles. 
Each encoder consists of deep convolutional architecture as described in subsection \ref{impl_details}.
We elegantly exploit the inherent multi-scale pyramidal structure of features at multiple layers of deep neural networks. We consider the features from the final three convolutional layers, conv\_6, conv\_7 and conv\_8 layers. These features are aggregated and followed by a fully connected layer to obtain the feature representation for images in each view.

The encoders for aerial and street-view images do not share the weights. Since the cross-view images are captured from different viewpoints, the visual entities exhibit drastic domain changes. 
The two encoders operate on these sets of diverse images, so it is understandable that the weight sharing is not a good choice.
On the other hand, the encoders for $I_{a'}$ and $I_a$ share the weights, since both images represent the aerial domain. 
This way, the aerial encoders learn weights suitable for the synthesized image $I_{a'}$ as well as the real image $I_a$. 
Thus, $f_{a'}$ effectively forces the features $f_a$ to be closer to $f_g$ and bridges the domain gap between the two views. 
This is possible because the transformed image $I_{a'}$  captures representations of $I_g$ which are easier for the network to learn from $I_{a'}$ than it would be when learning directly from $I_g$.

 This strategy leverages the synthesized images at training time, but does not require them during the testing. The auxiliary loss between $I_{a'}$ and $I_a$ influences the aerial image encoder to learn representations for aerial images by considering the synthesized aerial image. We train our network jointly on these image triads ($I_g$, $I_{a'}$ and $I_a$) using weighted soft-margin ranking loss \cite{Hu_2018_CVPR}, which is explained next.

\subsubsection{Weighted Soft-margin Triplet Loss}

\hspace{10pt} Consider a feature embedding $f_{g}$ of ground-level image, $f_{a-pos}$ of the corresponding matching aerial image and a non-matching aerial image feature $f_{a-neg}$.  The triplet loss \cite{DBLP:journals/corr/HermansBL17} aims to bring the matching feature $f_{a-pos}$ closer to $f_{g}$ while at the same time pushes $f_{a-neg}$ away. Here, if $d_{p}$ is the Euclidean distance between positive samples ($f_{g}$, $f_{a-pos}$) and $d_{n}$ is the Euclidean distance between negative/non-matching samples ($f_{g}$, $f_{a-neg}$), we try to minimize $d_{p}$ as well as maximize $d_{n}$. The triplet loss is expressed as shown below:

\vspace{-10pt}
\begin{equation}\label{triplet} 
L_{triplet} = max( 0, m + d_{p} - d_{n}),
\end{equation}

\noindent where, $m$ is a margin that specifies a minimum distance between non-matching pairs. 

In order to avoid the necessity of explicitly deciding the margin for triplet loss, soft-margin triplet loss is popular and is expressed as given in Equation \ref{triplet-soft} below:

\vspace{-10pt}

\begin{equation}\label{triplet-soft}
 L_{soft} = ln( 1 + e^d),
\end{equation}

\vspace{-5pt}

\noindent where $d = d_{p} - d_{n}$.

In our work, we use the weighted soft margin triplet loss \cite{Hu_2018_CVPR} as given in Equation \ref{weighted-soft}:

\vspace{-10pt}
\begin{equation}\label{weighted-soft}
 L_{weighted} = ln( 1 + e^{\alpha d}).
\end{equation}
We use  $ \alpha $ = 10, which  results in better convergence than $ \alpha $ = 1.

We incorporate the auxiliary loss between the synthesized aerial images, $I_a'$, and the real aerial images, $I_a$,  along with the loss between the real aerial, $I_a$,  and the ground images, $I_g$, for joint feature learning using the Equation \ref{joint-loss} below: 
\begin{equation}\label{joint-loss}
L_{joint} = \lambda_1 L_{weighted}{(I_g, I_a)} 
+  \lambda_2 L_{weighted}{(I_{a'}, I_a).}
\end{equation}

Here, $\lambda_1$ and $\lambda_2$ are balancing factors between the losses for ($I_g$, $I_a$) and ($I_{a'}$, $I_a$) pairs respectively.



\subsection{Feature Fusion} 

In the above method, the synthesized aerial image is used during the training only, for bridging the domain gap between the real aerial and ground view images; but is neglected during testing. Since the features of the synthesized image contain complementary information that assist in joint feature learning, we attempt to further exploit them. We fuse the ground image features $f_g$ with synthesized aerial image features $f_{a'}$ and find a robust representation $f_{g^*}$ for the query ground image. 

The fusion architecture is shown in Figure \ref{fig:fusion-net}. We use the trained joint feature learning network as feature extractor for our feature fusion task. We first concatenate the features from ground query image with the features from synthesized aerial image. The concatenated features need to be refined to obtain a generalized representation for query image $f_{g^*}$. We achieve this by passing through a fully-connected layer in the upper stream. 
The features $f_a$ from the lower stream need to be optimized against the refined features from upper fully-connected layer. So, we add a fully-connected layer in the lower stream that learns the generalized representations,  $f_{a^*}$, for the aerial images. During the testing, the fused feature representation $f_{g^*}$ for query image $I_g$ is compared against the representations $f_{a^*}$ for aerial images for image matching.

\begin{figure*}[t]
\begin{center}
   \includegraphics[width=0.92\linewidth]{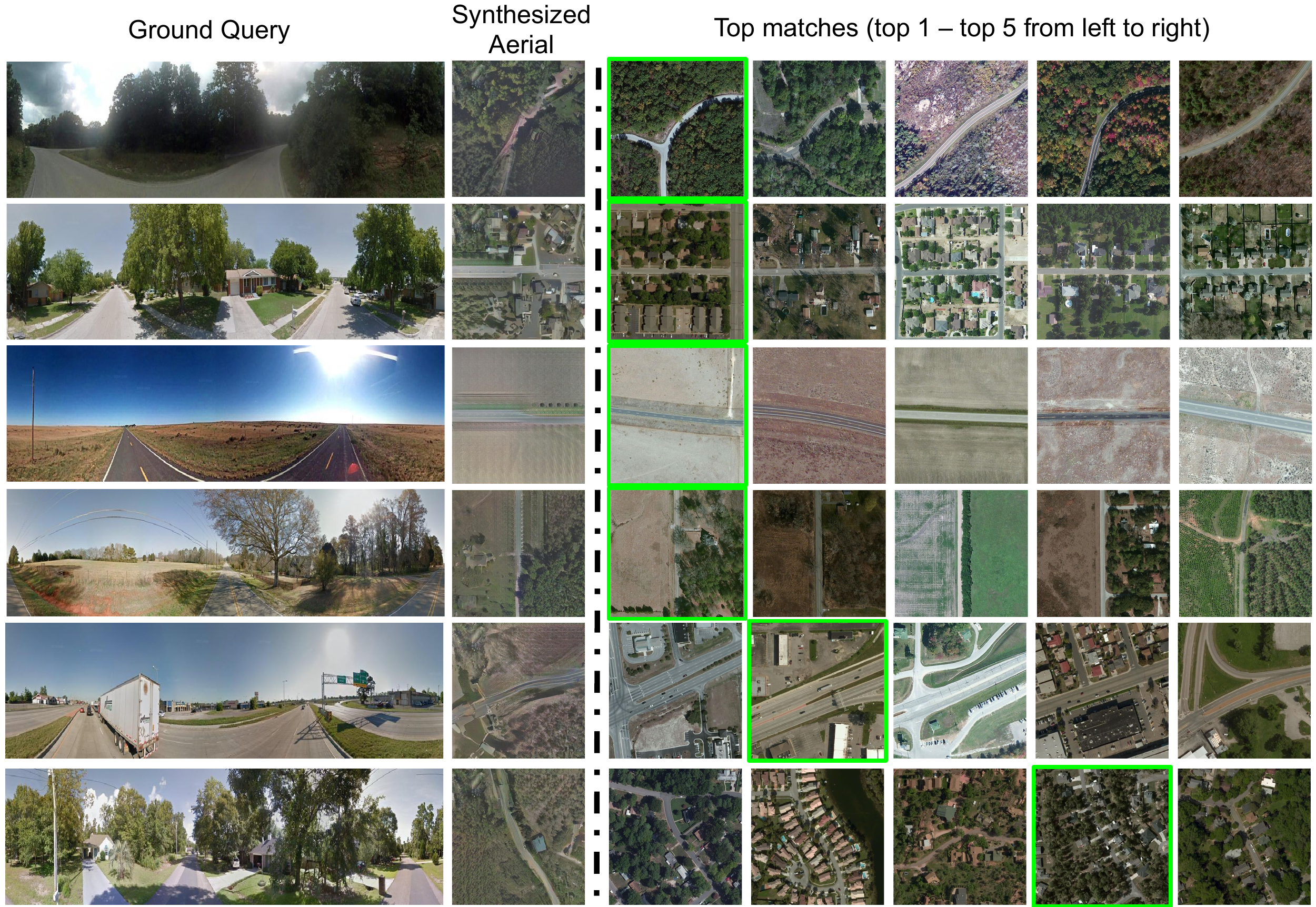}
  \vspace{-10pt}
\end{center}
   \caption{\small Image retrieval examples on CVUSA dataset \cite{zhai2017crossview}. For each query ground-level panorama, the synthesized aerial image is shown alongside, followed by the five closest aerial images retrieved by proposed Feature Fusion method. The correct matching (ground truth) aerial images are shown in green boxes. Rows 5 and 6 show examples where the ground truth aerial images are retrieved at the second and fourth positions respectively.
}
\label{fig:qual_cvusa}

\vspace{-10pt}

\end{figure*}


\section{Experimental Setup} 
This section deals with the datasets we used and the experimental setups we followed in our work. 
\subsection{Datasets} We conduct experiments on CVUSA dataset \cite{zhai2017crossview} to compare our work with existing methods. We also collect a new dataset, OP dataset,  from urban areas of Orlando and Pittsburgh with geo-information. The other benchmark dataset, GT-Crossview \cite{Vo2016} doesn't contain the ground level panorama, thus making it infeasible to synthesize meaningful aerial image. Also, the GT-Crossview dataset has aligned image pairs in training set, whereas unaligned image pairs in test set with no direction information, so the synthesized aerial images for test case will be randomly oriented relative to aerial images in the reference database, thus it is not possible to use this dataset in our framework.

\noindent \textbf{CVUSA: } CVUSA is a benchmark dataset for cross-view image matching with 35,532 satellite and ground-panorama image pairs for training and 8,884 pairs for testing. Aerial images are $750\times 750$ and ground-panorama are $224 \times 1232$ in resolutions.
Sample images from this dataset are shown in Figure \ref{fig:qual_cvusa}. 

\noindent \textbf{Orlando-Pittsburgh (OP) dataset: } We collect image pairs from two US cities, Orlando and Pittsburgh with ground-truth geo-locations. We call it Orlando-Pittsburgh (OP) dataset. The dataset covers urban areas of the cities, entirely different from the rural areas in CVUSA dataset. Figure \ref{fig:qual_OP} shows some example images of this dataset. The dataset contains 1,910 training and 722 testing pairs of aerial and ground-panorama images. The resolutions are $640 \times 640$ for aerial images and $416 \times 832$ for panoramas. Primary motivation to collect this dataset is to evaluate the generalization of the proposed methods in urban locations and to compute matching accuracy in terms of distance (meters); and the unavailability of such datasets publicly. Though small-scale, this dataset should be useful for future research in this direction. 

%

\subsection{Implementation Details} \label{impl_details}
We present the implementation details of our cross-view synthesis network and the proposed image matching networks in this section.

\vspace{5pt}
\noindent \textbf{Cross-View Synthesis network: } 
The generator of cross-view synthesis network, shown as \textit{Generator} in Figures \ref{fig:intro_new} and \ref{fig:nw_arch} has an encoder and two decoders, similar to the X-Fork architecture in \cite{Regmi_2018_CVPR}. The input to the encoder is a 4-channel image; 3-RGB channels and an edgemap, stacked together. The decoders generate cross-view image and its segmentation map, for a given input. The network consists of blocks of Convolution, Batch Normalization and Leaky ReLU layers. Convolutional kernels of size 
$4 \times 4$ with a stride of 2 are used that downsamples the feature maps after each convolution, and to upsample the feature maps after each upconvolution operation. 
We reshape the features at bottleneck to adjust the feature shape and pass through the decoders.
The six blocks of decoders share the weights whereas the final two blocks don't. 
The discriminator network has similar architecture to the one used in \cite{Regmi_2018_CVPR}.
We train the GAN end-to-end using Torch \cite{Collobert_NIPSWORKSHOP_2011} implementation. 
The weights are initialized with a random Gaussian distribution with zero mean and 0.02 standard deviation.


\vspace{5pt}
\noindent \textbf{Joint Feature Learning  network: } 
Each stream (encoder) 
of joint feature learning network in Figure \ref{fig:aux-loss} consists of seven convolutional layers, each followed by ReLU activations. Dropouts are applied after the final three ReLU layers. The features after these dropouts are flattened and then concatenated to obtain multi-scale representation of the input image. This is followed by a fully-connected layer for dimensionality reduction to obtain 1,000-dimensional feature vector for each input. The two-stream baselines are trained from scratch with Xavier initialization. The joint feature learning network is initialized with weights from the two-stream network trained on ($I_g$, $I_a$) image pairs and the loss function is optimized as shown in Equation \ref{joint-loss}. We use $\lambda_1$ = 10 and $\lambda_2$ = 1, weighing more on the loss term for ($I_g$, $I_a$) pairs because of their superior performance over ($I_{a'}$, $I_a$) in image matching as reported in Table \ref{tab:one_ten_1percent} and objectively $I_a'$ is used as an auxiliary information, only during the training in joint feature learning network. 

\vspace{5pt}
\noindent \textbf{Feature Fusion network} 
The Feature Fusion network in Figure \ref{fig:fusion-net} has two fully-connected layers, one each for aerial and ground feature branches. The upper FC layer takes 2000-dimensional fused feature and translates it to a 1000-dimensional feature representation.  The input to the lower FC layer is $f_a$ that is mapped to a 1000 dimensional feature representation. The FC layers are randomly initialized with a uniform distribution with zero mean and 0.005 standard deviation. 

The two-stream baselines and the proposed joint feature learning and feature fusion networks are implemented using Tensorflow \cite{45381} with Adam optimizer (lr = $10^{-5}$) and dropout = 0.5. A batch size of B = 30 for experiments on two-stream networks and B = 24 for joint feature learning networks is used. Weighted soft-margin triplet loss  is used for training in all the experiments. An exhaustive mini-batch strategy \cite{Vo2016} is employed to maximize the number of triplets within each batch.
For each image in a batch of B images, we have 1 positive pair and (B-1) negative pairs for each ground image, and (B-1) negative pairs for each aerial image. So, for B images, we have B positive pairs and 2 x B x (B-1) negative pairs.
 Further training is continued with in-batch hard negative mining; by training each positive pair against the most negative sample (i.e. smallest distance) in the batch. Code and dataset is publicly available
\footnote{https://github.com/kregmi/cross-view-image-matching}.


In summary, GAN  is first trained to generate the cross-view image $I_{a'}$ for the ground panorama $I_g$. Next,  the synthesized images are used for joint feature learning in our proposed method.

\section{Results} 

We present an extensive analysis of our proposed method demonstrating the effectiveness of synthesized images for image retrieval to bridge the domain gap between the cross-view images. We also provide the comparison of our work with the state-of-the-art methods on the CVUSA dataset. Finally, we present an evaluation on geo-localization task on the OP dataset.

\subsection{Evaluation Metric}
The common metric for evaluation of image based matching task is to compute the recall accuracy. A matching is successful for a query street-view image if the correct match lies within a set of closest images in Euclidean distance of the representative features. We report top-1\% accuracy for ease of comparison with previous works. We also report top-1 and top-10 recalls on CVUSA dataset. 


\begin{table}[t]
 \small
  \centering
  \renewcommand{\tabcolsep}{1.2mm}  
  \caption{\small Comparison of Top-1, Top-10 and Top-1\% recall for the baselines and the proposed approaches (first panel) and with previous methods (second panel) on CVUSA Dataset \cite{zhai2017crossview}.}
  \vspace{-5pt}
  \label{tab:one_ten_1percent}
    \begin{tabular}{l|ccc}
        \toprule
        \textbf{Method} & \textbf{Top-1} &\textbf{Top-10} & \textbf{Top-1\%} \\
        \midrule
        \midrule

        Two-stream baseline ($I_{a'}$, $I_a$) & $ 10.23 \% $& $ 35.10 \% $  & $ 72.58 \% $\\

         Two-stream baseline ($I_g$, $I_a$) & 18.45\%  &$ 48.98\%  $ & 82.94\%   \\
    
        Joint Feat. Learning ($I_{a'}$, $I_a$) & 14.31\%  & 48.75\%  & 86.47\%    \\
        
        Joint Feat. Learning ($I_g$, $I_a$) & 29.75\%  & 66.34\%  & 92.09\%   \\

        Feature Fusion & {\bf 48.75\%}  & {\bf 81.27\%}  & \textbf{95.98\% }   \\
    
            \midrule

        Workman et al. \cite{workman2015wide} & - & - & 34.3\% \\
        
        Zhai et al. \cite{zhai2017crossview}  & - & - & 43.2\%  \\
        
        Vo and Hays \cite{Vo2016} & - & - & $ 63.7\% $\\
        
        CVM-Net-I \cite{Hu_2018_CVPR} & 22.53\% & $ 63.28\% $ & 91.4\%   \\ 

        CVM-Net-II \cite{Hu_2018_CVPR} & 11.18\%  & 43.51\%  & 87.2\%   \\     
        \bottomrule
    \end{tabular}
    \vspace{-5pt}
\end{table}

\subsection{Results of Our Approach}
 We evaluate our model variants in terms of retrieval accuracy on the CVUSA dataset \cite{zhai2017crossview}. The results are reported in Table \ref{tab:one_ten_1percent} (first panel). 
 
\vspace{5pt}
\noindent \textbf{Baseline Comparison} (first and second rows in Table  \ref{tab:one_ten_1percent} (first panel)): The two-stream networks trained employing image pairs ($I_g$, $I_a$) and ($I_{a'}$, $I_a$), where first image in each tuple is the query, are the baselines.
 We observe that the synthesized image $I_{a'}$ as a query performs quite well with 72.58\% for top-1\% recall but slightly lower than $I_g$ as query (82.94\%). This means that the synthesized images capture fair amount of information from the ground panorama whereas they are not yet completely dependable for cross-view image retrieval and we need to consider real ground images as well. This provided us the motivation for joint feature learning.

\vspace{5pt}
\noindent \textbf{Joint Feature Learning} (third  and fourth rows in Table  \ref{tab:one_ten_1percent} (first panel)):  For joint feature learning, as explained earlier, image triads ($I_g$, $I_a$ and $I_{a'}$) are used during training and only ($I_g$, $I_a$) pairs are used during the testing. 
We report an improvement of about 9\% in top-1\% retrieval accuracy over two-stream baseline ($I_g$, $I_a$) by joint feature learning. 
The improvement suggests that the synthesized aerial images contain  features complementary to ground image features that facilitate the network to learn better representations for aerial images during the joint feature learning. 
The synthesized aerial image as an auxiliary information between the ground and aerial images is successful in forcing them to bring their feature representations closer to each other during the joint feature learning. 

Since the representations for $I_g$, $I_a$ and $I_{a'}$ were learned together during joint feature learning, we were curious to evaluate how well the feature representations for $I_{a'}$ perform in image matching. Unsurprisingly, we obtain an improvement of about 14\% in top-1\% retrieval accuracy over two-stream baseline ($I_{a'}$, $I_{a}$). This improvement further consolidated the belief that the learned features for $I_g$ and $I_{a'}$ are complementary to each other and can be fused together to obtain robust descriptor for the ground image.

\vspace{5pt}

\noindent \textbf{Feature Fusion:} (fifth row in Table  \ref{tab:one_ten_1percent} (first panel)): The Feature Fusion approach fuses the synthesized image features with the ground image features to obtain a representative feature for the query. This provides further improvement of 3.89\% in top-1\% accuracy (compare fourth and fifth rows). The significance of feature fusion can be measured by about 19\% improvement in top-1 retrieval accuracy over joint feature learning. This improvement further signifies that the synthesized image features are complementary to street-view image features that should be exploited to obtain better features for cross-view matching. The qualitative results are shown in Figure \ref{fig:qual_cvusa}. The query ground images and the synthesized aerial images along with five closest images are shown in each row.


\subsection{Comparison to Existing Methods}
\begin{figure}[t]
\begin{center}
   \includegraphics[width=0.99\linewidth]{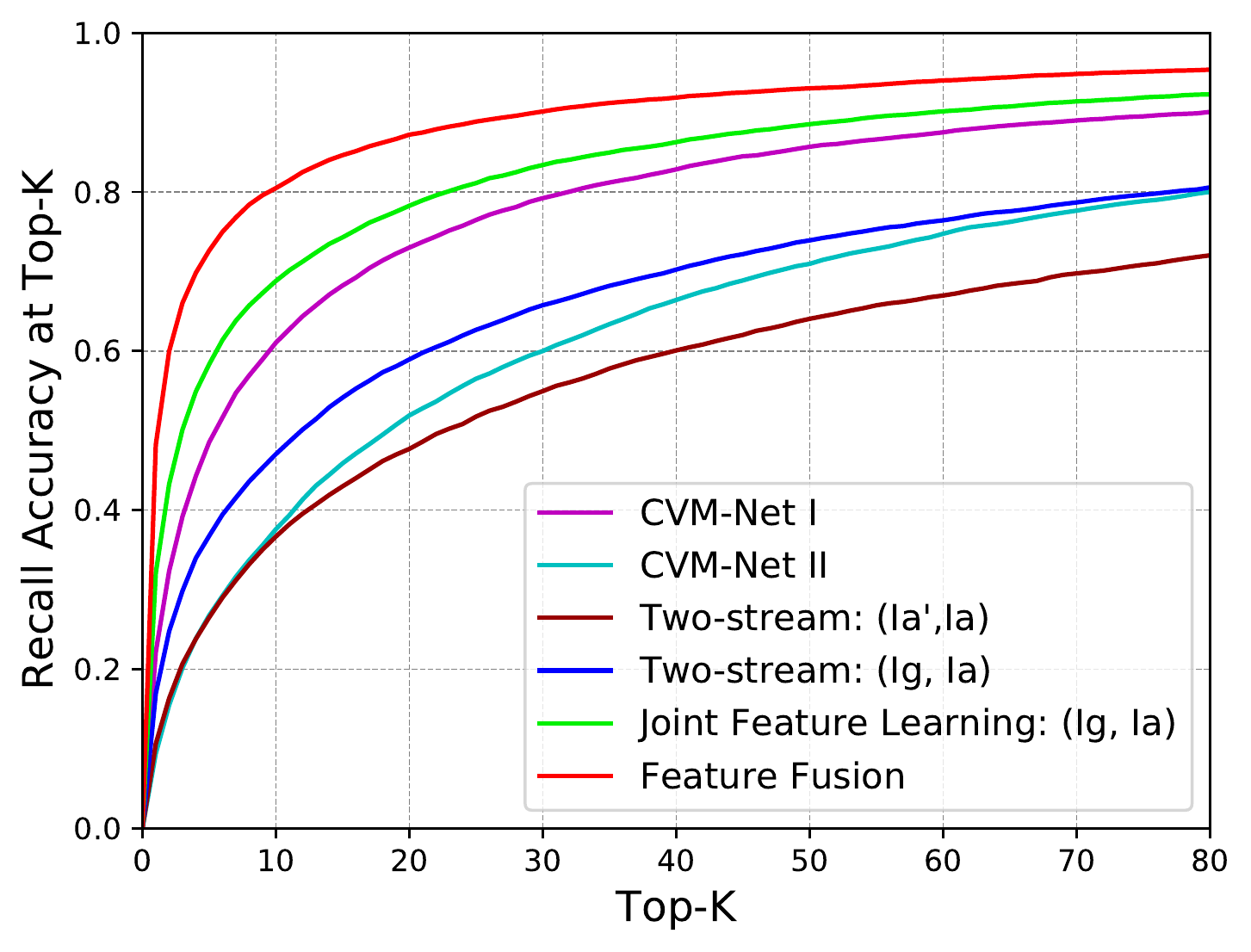}
   \vspace{-20pt}
\end{center}
   \caption{\small Comparison of different versions of our methods with CVM-Net I and CVM-Net II \cite{Hu_2018_CVPR} on CVUSA dataset \cite{zhai2017crossview}. 
   }
\label{fig:auc}
\end{figure}

We compare our work with the previous approaches by \cite{workman2015wide, zhai2017crossview, Vo2016, Hu_2018_CVPR} on CVUSA dataset \cite{zhai2017crossview}. We report the top-1, top-10 and top1-\% accuracies for state-of-the-art CVM-Net \cite{Hu_2018_CVPR} and our methods. The results are shown in Table \ref{tab:one_ten_1percent} (second panel). We observe that the Joint Feature Learning outperforms ( fourth row in Table  \ref{tab:one_ten_1percent} (first panel)) the previous works and is further boosted by Feature Fusion ( fifth row in Table  \ref{tab:one_ten_1percent} (first panel)). We achieve an overall 4.58\% improvement over SOTA CVM-Net \cite{Hu_2018_CVPR} for top-1\% recall accuracy. We obtain significant increments of more than 26\% and 18\% in top-1 and top-10 accuracies over CVM-Net-I \cite{Hu_2018_CVPR}.
We also plot top-K recall accuracy for K = 1 to 80 for our methods as compared with previous approaches in Figure \ref{fig:auc}. It illustrates that various versions of our proposed method outperform the existing state-of-the-art approaches for all values of $K$. 


\begin{table}[]
 \small
  \centering
  \renewcommand{\tabcolsep}{1.4mm}  
  
  \caption{\small Ablation Study on CVUSA Dataset \cite{zhai2017crossview}. The reported numbers are the retrieval accuracies for Feature Fusion network for specified ablation criteria. }
  \vspace{-5pt}
  \label{tab:ablation}
    \begin{tabular}{l|ccc}
        \toprule
        \textbf{Ablation Criteria} & \textbf{Top-1} &\textbf{Top-10} & \textbf{Top-1\%} \\
        \midrule
        \midrule

        Single Scale Features & $ 8.01 $ \% & $ 32.62 $ \%  & $ 74.41 $\% \\
  
        Global Avg. Pooling (GAP) & $ 16.13 \% $ & $ 51.72 \% $ & $ 87.68 \% $\\
        Weight Sharing & $ 29.94 \% $ & $ 68.24 \% $ & $ 93.42 \% $\\
        
        \midrule
        Multi-scale  Features  & & &   \\
          + No GAP + No Wt. Sharing & {\bf 48.75\%}  & {\bf 81.27\%}  & \textbf{95.98\% } 
\\
        \bottomrule
  
    \end{tabular}
\end{table}

\subsection{Ablation Study} 
We conduct the following ablation studies to understand the impact of different choices made in the proposed networks. For the experiments on ablation, the joint feature learning and feature fusion networks are used with specified setups: a) single scale features - only the final layer features are matched, b) global average pooling (GAP) - GAP operation suppresses the spatial dimension of feature maps, substantially reducing the feature size, and c) weight sharing between the encoders for aerial and ground images. All these methods reduce the number of parameters used in the network. 

\vspace{5pt}
\noindent \textbf{Single Scale vs. Multi-scale Features:} For this ablation, joint feature learning network with single scale features is trained first followed by experiments using the Feature Fusion network. 
The features after the final convolutional block (conv\_8) are considered as single scale features. These are the representative features for the given input image and are used for matching. We do not employ global average pooling and weight sharing in this ablation for direct comparison of the single-scale vs. multi-scale feature representations.  
The scores are reported in Table \ref{tab:ablation} (first row for single scale and fourth row for multi-scale features). The results signify that features from conv\_6 and conv\_7 are also crucial in image matching rather than just using the features from final conv\_8 layer only. 
The results demonstrate the importance of aggregating the multi-scale features for cross-view matching task.

\vspace{5pt}
\noindent \textbf{Pooling vs. No Pooling: }
We also conduct ablations on using global average pooling \cite{lin2013network} in our experiments. Global average pooling is a popular approach to reduce the spatial dimensions of the features and consequently reduce the number of parameters in the network. We experimented with using global average pooling layer before concatenating the features from multiple scales. The result is reported in Table \ref{tab:ablation} (second row for using GAP and fourth row without using GAP, rest of the architecture being the same). We observe that the loss of spatial information in features severely impacts the retrieval performance. 

\vspace{5pt}
\noindent \textbf{Weight Sharing vs. No Weight Sharing:} We believe that the two branches receiving the input images from completely different viewpoints as is the case with aerial and ground -view images should not share the weights. Even though the networks will be looking at same scene contents their representations from the two views are drastically different, thus suggesting that the networks should freely evolve their weights based on the input they receive. The results are  reported in Table \ref{tab:ablation} (third row for weight sharing and fourth row for without weight sharing, remainder of the setup being the same). The numbers clearly suggest that no weight sharing is fairly an easy choice, especially looking at the difference of about 18\% in Top-1 accuracies. 

\begin{figure}[t]
\begin{center}
   \includegraphics[width=0.86\linewidth]{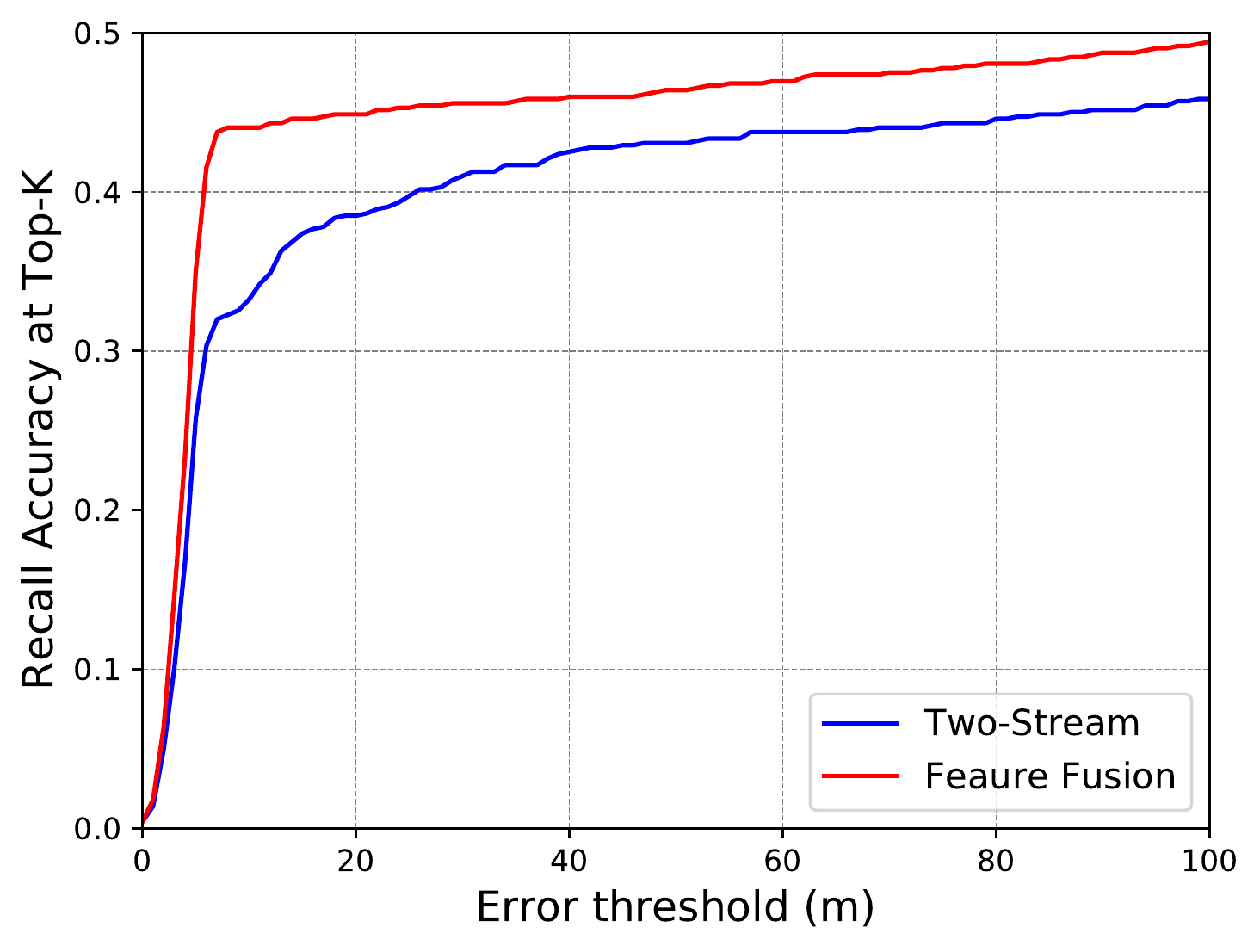}
  \vspace{-20pt}
\end{center}
   \caption{\small Geo-localization results on the OP dataset with different error thresholds.}
\label{fig:gps_retrieval}
\vspace{-15pt}
\end{figure}


\begin{figure}[t]
\begin{center}
   \includegraphics[width=0.99\linewidth]{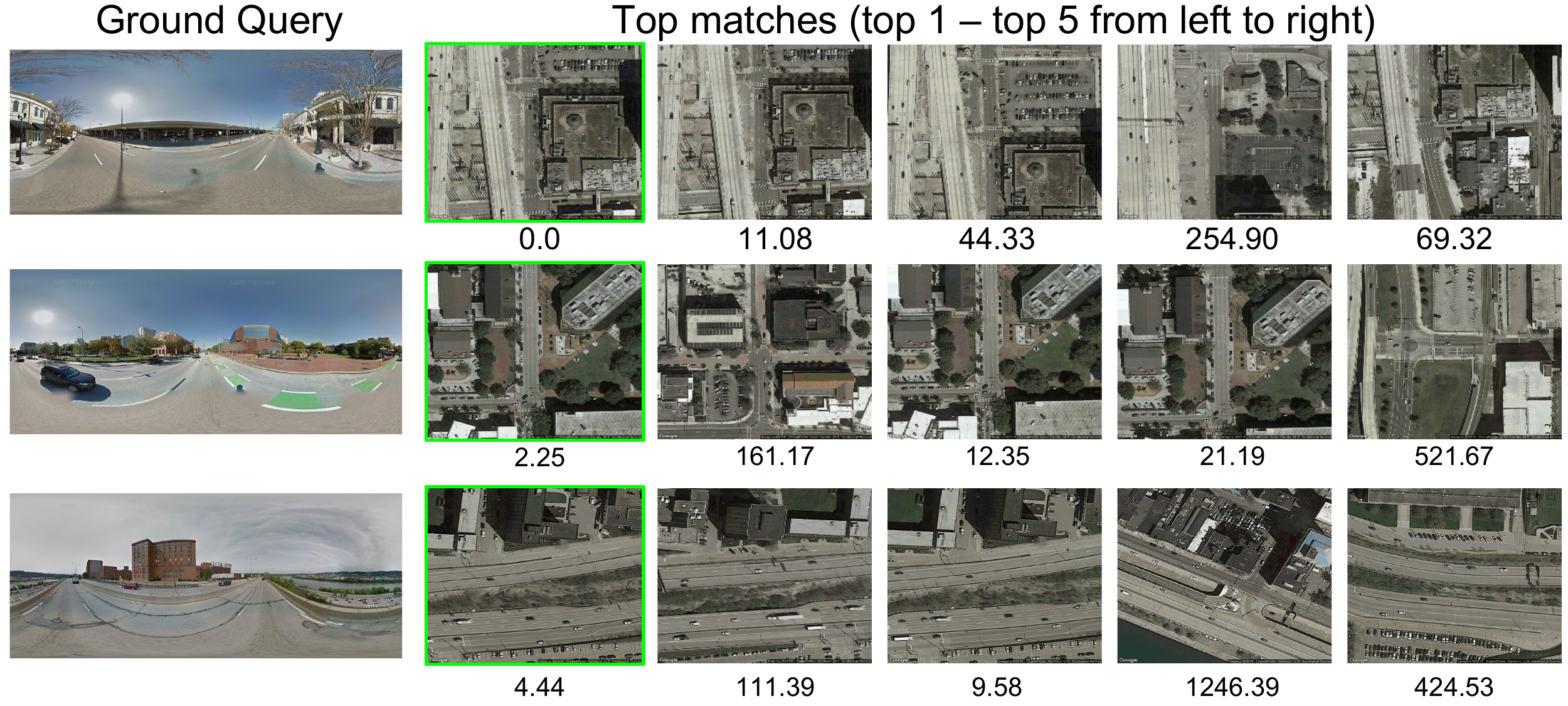}
  \vspace{-15pt}
\end{center}
   \caption{\small Image retrieval examples on the OP dataset. The correct aerial image matches are shown in green borders. 
   The numbers below each aerial image shows its distance in meters from query ground image.
}
\label{fig:qual_OP}
\vspace{-5pt}
\end{figure}


\subsection{Cross-view Localization} We use the Orlando-Pittsburgh (OP) dataset for image based geo-localization. We want to determine the gps location of the query image by assigning it the location of closest retrieved aerial image. The query image is correctly geo-localized if it is located within a threshold distance in meters from its ground truth position. 

The recall accuracy with respect to distance threshold in meters is plotted in Figure \ref{fig:gps_retrieval}. We observe that our proposed Feature Fusion method can retrieve images close to its geo-location with higher accuracy than the baseline which can be attributed to its superiority in Top-1 recall. 


The image retrieval examples on the OP dataset are shown in Figure \ref{fig:qual_OP}. The ground query images are followed by the five closest aerial images. 
Even though the retrieved images are very similar to each other, we are able to retrieve the correct match at Top-1 position. 
The Top-1 recall is reported in Table \ref{tab:top1_OP}. The results affirm that the proposed methods are generalizable to urban cities of OP dataset as well as rural areas of CVUSA dataset. 


\begin{table}[]
 \small
  \centering
  \renewcommand{\arraystretch}{.92}
  \renewcommand{\tabcolsep}{.45mm}  
  
  \caption{\small Top-1 retrieval accuracy on Orlando-Pittsburgh Dataset.}
  \vspace{-5pt}
  \label{tab:top1_OP}
    \begin{tabular}{c|c|c}
        \toprule
        \textbf{Two-stream ($I_g$, $I_a$)} & \textbf{Joint Feat. Learning} &\textbf{Feature Fusion}  \\
        \midrule
        \midrule
       $ 30.61\% $ & $ 38.36\% $ & $ 45.57 \% $ \\
        \bottomrule
    \end{tabular}
    \vspace{-5pt}
\end{table}


\section{Conclusion} 
In this paper, we have presented a novel and practical approach to cross-view image retrieval by transforming the query image to target view to obtain a better scene understanding. We showed that the synthesized aerial images can be seamlessly incorporated in cross-view matching pipeline by joint feature training to bridge the domain gap between the aerial and street-view images. Also, the ground image features and the corresponding synthesized aerial image features are fused to obtain a robust descriptor of the ground image. We obtained significant improvements over state-of-the-art methods on the challenging CVUSA dataset. \\    

\noindent \textbf{Acknowledgements.} The authors would like to thank Chen Chen and Yonatan Tariku for helpful discussions; Vijay Gunasekaran and Hassan Maqbool for their contribution in OP Dataset collection and the anonymous reviewers for the critical feedback.


{\small
\bibliographystyle{ieee_fullname}
\bibliography{references}

\begin{thebibliography}{10}\itemsep=-1pt

\bibitem{45381}
Martin Abadi, Paul Barham, Jianmin Chen, Zhifeng Chen, Andy Davis, Jeffrey
  Dean, Matthieu Devin, Sanjay Ghemawat, Geoffrey Irving, Michael Isard,
  Manjunath Kudlur, Josh Levenberg, Rajat Monga, Sherry Moore, Derek~G. Murray,
  Benoit Steiner, Paul Tucker, Vijay Vasudevan, Pete Warden, Martin Wicke, Yuan
  Yu, and Xiaoqiang Zheng.
\newblock Tensorflow: A system for large-scale machine learning.
\newblock In {\em 12th USENIX Symposium on Operating Systems Design and
  Implementation (OSDI 16)}, pages 265--283, 2016.

\bibitem{almahairi2018augmented}
Amjad Almahairi, Sai Rajeshwar, Alessandro Sordoni, Philip Bachman, and Aaron
  Courville.
\newblock Augmented cyclegan: Learning many-to-many mappings from unpaired
  data.
\newblock In {\em International Conference on Machine Learning}, pages
  195--204, 2018.

\bibitem{arandjelovic2016netvlad}
Relja Arandjelovic, Petr Gronat, Akihiko Torii, Tomas Pajdla, and Josef Sivic.
\newblock Netvlad: Cnn architecture for weakly supervised place recognition.
\newblock In {\em Proceedings of the IEEE Conference on Computer Vision and
  Pattern Recognition}, pages 5297--5307, 2016.

\bibitem{10.1007/978-3-642-33863-2_18}
Mayank Bansal, Kostas Daniilidis, and Harpreet Sawhney.
\newblock Ultra-wide baseline facade matching for geo-localization.
\newblock In Andrea Fusiello, Vittorio Murino, and Rita Cucchiara, editors,
  {\em Computer Vision -- ECCV 2012. Workshops and Demonstrations}, pages
  175--186, Berlin, Heidelberg, 2012. Springer Berlin Heidelberg.

\bibitem{bell2016inside}
Sean Bell, C Lawrence~Zitnick, Kavita Bala, and Ross Girshick.
\newblock Inside-outside net: Detecting objects in context with skip pooling
  and recurrent neural networks.
\newblock In {\em Proceedings of the IEEE conference on computer vision and
  pattern recognition}, pages 2874--2883, 2016.

\bibitem{canny1987computational}
John Canny.
\newblock A computational approach to edge detection.
\newblock In {\em Readings in computer vision}, pages 184--203. Elsevier, 1987.

\bibitem{Choi_2018_CVPR}
Yunjey Choi, Minje Choi, Munyoung Kim, Jung-Woo Ha, Sunghun Kim, and Jaegul
  Choo.
\newblock Stargan: Unified generative adversarial networks for multi-domain
  image-to-image translation.
\newblock In {\em The IEEE Conference on Computer Vision and Pattern
  Recognition (CVPR)}, June 2018.

\bibitem{Collobert_NIPSWORKSHOP_2011}
Ronan Collobert, Koray Kavukcuoglu, and Cl{\'{e}}ment Farabet.
\newblock Torch7: A matlab-like environment for machine learning.
\newblock In {\em BigLearn, NIPS Workshop}, 2011.

\bibitem{Deng:2018:LDG:3274895.3274969}
Xueqing Deng, Yi Zhu, and Shawn Newsam.
\newblock What is it like down there?: Generating dense ground-level views and
  image features from overhead imagery using conditional generative adversarial
  networks.
\newblock In {\em Proceedings of the 26th ACM SIGSPATIAL International
  Conference on Advances in Geographic Information Systems}, SIGSPATIAL '18,
  pages 43--52, New York, NY, USA, 2018. ACM.

\bibitem{elfeki2018third}
Mohamed Elfeki, Krishna Regmi, Shervin Ardeshir, and Ali Borji.
\newblock From third person to first person: Dataset and baselines for
  synthesis and retrieval.
\newblock {\em arXiv preprint arXiv:1812.00104}, 2018.

\bibitem{ghouaiel2016coupling}
Nehla Ghouaiel and S{\'e}bastien Lef{\`e}vre.
\newblock Coupling ground-level panoramas and aerial imagery for change
  detection.
\newblock {\em Geo-spatial Information Science}, 19(3):222--232, 2016.

\bibitem{goodfellow2014generative}
Ian Goodfellow, Jean Pouget-Abadie, Mehdi Mirza, Bing Xu, David Warde-Farley,
  Sherjil Ozair, Aaron Courville, and Yoshua Bengio.
\newblock Generative adversarial nets.
\newblock In {\em Advances in neural information processing systems}, pages
  2672--2680, 2014.

\bibitem{hariharan2015hypercolumns}
Bharath Hariharan, Pablo Arbel{\'a}ez, Ross Girshick, and Jitendra Malik.
\newblock Hypercolumns for object segmentation and fine-grained localization.
\newblock In {\em Proceedings of the IEEE conference on computer vision and
  pattern recognition}, pages 447--456, 2015.

\bibitem{Hays:2008:im2gps}
James Hays and Alexei~A. Efros.
\newblock im2gps: estimating geographic information from a single image.
\newblock In {\em Proceedings of the {IEEE} Conf. on Computer Vision and
  Pattern Recognition ({CVPR})}, 2008.

\bibitem{Hays2015LargeScaleIG}
James Hays and Alexei~A. Efros.
\newblock Large-scale image geolocalization.
\newblock In {\em Multimodal Location Estimation of Videos and Images}, 2015.

\bibitem{DBLP:journals/corr/HermansBL17}
Alexander Hermans, Lucas Beyer, and Bastian Leibe.
\newblock In defense of the triplet loss for person re-identification.
\newblock {\em CoRR}, abs/1703.07737, 2017.

\bibitem{honari2016recombinator}
Sina Honari, Jason Yosinski, Pascal Vincent, and Christopher Pal.
\newblock Recombinator networks: Learning coarse-to-fine feature aggregation.
\newblock In {\em Proceedings of the IEEE Conference on Computer Vision and
  Pattern Recognition}, pages 5743--5752, 2016.

\bibitem{Hu_2018_CVPR}
Sixing Hu, Mengdan Feng, Rang M.~H. Nguyen, and Gim Hee~Lee.
\newblock Cvm-net: Cross-view matching network for image-based ground-to-aerial
  geo-localization.
\newblock In {\em The IEEE Conference on Computer Vision and Pattern
  Recognition (CVPR)}, June 2018.

\bibitem{pix2pix2017}
Phillip Isola, Jun-Yan Zhu, Tinghui Zhou, and Alexei~A Efros.
\newblock Image-to-image translation with conditional adversarial networks.
\newblock {\em CVPR}, 2017.

\bibitem{pmlr-v70-kim17a}
Taeksoo Kim, Moonsu Cha, Hyunsoo Kim, Jung~Kwon Lee, and Jiwon Kim.
\newblock Learning to discover cross-domain relations with generative
  adversarial networks.
\newblock In Doina Precup and Yee~Whye Teh, editors, {\em Proceedings of the
  34th International Conference on Machine Learning}, volume~70 of {\em
  Proceedings of Machine Learning Research}, pages 1857--1865, International
  Convention Centre, Sydney, Australia, 06--11 Aug 2017. PMLR.

\bibitem{kong2016hypernet}
Tao Kong, Anbang Yao, Yurong Chen, and Fuchun Sun.
\newblock Hypernet: Towards accurate region proposal generation and joint
  object detection.
\newblock In {\em Proceedings of the IEEE conference on computer vision and
  pattern recognition}, pages 845--853, 2016.

\bibitem{lin2017refinenet}
Guosheng Lin, Anton Milan, Chunhua Shen, and Ian Reid.
\newblock Refinenet: Multi-path refinement networks for high-resolution
  semantic segmentation.
\newblock In {\em Proceedings of the IEEE conference on computer vision and
  pattern recognition}, pages 1925--1934, 2017.

\bibitem{lin2013network}
Min Lin, Qiang Chen, and Shuicheng Yan.
\newblock Network in network.
\newblock {\em arXiv preprint arXiv:1312.4400}, 2013.

\bibitem{Lin_2013_CVPR}
Tsung-Yi Lin, Serge Belongie, and James Hays.
\newblock Cross-view image geolocalization.
\newblock In {\em The IEEE Conference on Computer Vision and Pattern
  Recognition (CVPR)}, June 2013.

\bibitem{lin2017feature}
Tsung-Yi Lin, Piotr Doll{\'a}r, Ross Girshick, Kaiming He, Bharath Hariharan,
  and Serge Belongie.
\newblock Feature pyramid networks for object detection.
\newblock In {\em Proceedings of the IEEE Conference on Computer Vision and
  Pattern Recognition}, pages 2117--2125, 2017.

\bibitem{long2015fully}
Jonathan Long, Evan Shelhamer, and Trevor Darrell.
\newblock Fully convolutional networks for semantic segmentation.
\newblock In {\em Proceedings of the IEEE conference on computer vision and
  pattern recognition}, pages 3431--3440, 2015.

\bibitem{maaten2008visualizing}
Laurens van~der Maaten and Geoffrey Hinton.
\newblock Visualizing data using t-sne.
\newblock {\em Journal of machine learning research}, 9(Nov):2579--2605, 2008.

\bibitem{DBLP:journals/corr/MirzaO14}
Mehdi Mirza and Simon Osindero.
\newblock Conditional generative adversarial nets.
\newblock {\em CoRR}, abs/1411.1784, 2014.

\bibitem{newell2016stacked}
Alejandro Newell, Kaiyu Yang, and Jia Deng.
\newblock Stacked hourglass networks for human pose estimation.
\newblock In {\em European Conference on Computer Vision}, pages 483--499.
  Springer, 2016.

\bibitem{radford2015unsupervised}
Alec Radford, Luke Metz, and Soumith Chintala.
\newblock Unsupervised representation learning with deep convolutional
  generative adversarial networks.
\newblock {\em arXiv preprint arXiv:1511.06434}, 2015.

\bibitem{pmlr-v48-reed16}
Scott Reed, Zeynep Akata, Xinchen Yan, Lajanugen Logeswaran, Bernt Schiele, and
  Honglak Lee.
\newblock Generative adversarial text to image synthesis.
\newblock In Maria~Florina Balcan and Kilian~Q. Weinberger, editors, {\em
  Proceedings of The 33rd International Conference on Machine Learning},
  volume~48 of {\em Proceedings of Machine Learning Research}, pages
  1060--1069, New York, New York, USA, 20--22 Jun 2016. PMLR.

\bibitem{Regmi_2018_CVPR}
Krishna Regmi and Ali Borji.
\newblock Cross-view image synthesis using conditional gans.
\newblock In {\em The IEEE Conference on Computer Vision and Pattern
  Recognition (CVPR)}, June 2018.

\bibitem{REGMI2019}
Krishna Regmi and Ali Borji.
\newblock Cross-view image synthesis using geometry-guided conditional gans.
\newblock {\em Computer Vision and Image Understanding}, 2019.

\bibitem{DBLP:journals/corr/RonnebergerFB15}
Olaf Ronneberger, Philipp Fischer, and Thomas Brox.
\newblock U-net: Convolutional networks for biomedical image segmentation.
\newblock {\em CoRR}, abs/1505.04597, 2015.

\bibitem{Sattler_2016_CVPR}
Torsten Sattler, Michal Havlena, Konrad Schindler, and Marc Pollefeys.
\newblock Large-scale location recognition and the geometric burstiness
  problem.
\newblock In {\em The IEEE Conference on Computer Vision and Pattern
  Recognition (CVPR)}, June 2016.

\bibitem{shan2014accurate}
Qi Shan, Changchang Wu, Brian Curless, Yasutaka Furukawa, Carlos Hernandez, and
  Steven~M Seitz.
\newblock Accurate geo-registration by ground-to-aerial image matching.
\newblock In {\em 2014 2nd International Conference on 3D Vision}, volume~1,
  pages 525--532. IEEE, 2014.

\bibitem{tian2017cross}
Yicong Tian, Chen Chen, and Mubarak Shah.
\newblock Cross-view image matching for geo-localization in urban environments.
\newblock In {\em Proceedings of the IEEE Conference on Computer Vision and
  Pattern Recognition}, pages 3608--3616, 2017.

\bibitem{torii2011visual}
Akihiko Torii, Josef Sivic, and Tomas Pajdla.
\newblock Visual localization by linear combination of image descriptors.
\newblock In {\em 2011 IEEE International Conference on Computer Vision
  Workshops (ICCV Workshops)}, pages 102--109. IEEE, 2011.

\bibitem{Vo2016}
Nam~N. Vo and James Hays.
\newblock {\em Localizing and Orienting Street Views Using Overhead Imagery},
  pages 494--509.
\newblock Springer International Publishing, Cham, 2016.

\bibitem{workman2015location}
Scott Workman and Nathan Jacobs.
\newblock On the location dependence of convolutional neural network features.
\newblock In {\em Proceedings of the IEEE Conference on Computer Vision and
  Pattern Recognition Workshops}, pages 70--78, 2015.

\bibitem{workman2015wide}
Scott Workman, Richard Souvenir, and Nathan Jacobs.
\newblock Wide-area image geolocalization with aerial reference imagery.
\newblock In {\em {IEEE International Conference on Computer Vision (ICCV)}},
  2015.

\bibitem{yi2017dualgan}
Zili Yi, Hao Zhang, Ping Tan, and Minglun Gong.
\newblock Dualgan: Unsupervised dual learning for image-to-image translation.
\newblock In {\em Proceedings of the IEEE International Conference on Computer
  Vision}, pages 2849--2857, 2017.

\bibitem{10.1007/978-3-642-15561-1_19}
Amir~Roshan Zamir and Mubarak Shah.
\newblock Accurate image localization based on google maps street view.
\newblock In Kostas Daniilidis, Petros Maragos, and Nikos Paragios, editors,
  {\em Computer Vision -- ECCV 2010}, pages 255--268, Berlin, Heidelberg, 2010.
  Springer Berlin Heidelberg.

\bibitem{DBLP:journals/pami/ZamirS14}
Amir~Roshan Zamir and Mubarak Shah.
\newblock Image geo-localization based on multiplenearest neighbor feature
  matching usinggeneralized graphs.
\newblock {\em {IEEE} Trans. Pattern Anal. Mach. Intell.}, 36(8):1546--1558,
  2014.

\bibitem{zemene2019large}
Eyasu Zemene, Yonatan~Tariku Tesfaye, Haroon Idrees, Andrea Prati, Marcello
  Pelillo, and Mubarak Shah.
\newblock Large-scale image geo-localization using dominant sets.
\newblock {\em IEEE transactions on pattern analysis and machine intelligence},
  41(1):148--161, 2019.

\bibitem{zhai2017crossview}
Menghua Zhai, Zachary Bessinger, Scott Workman, and Nathan Jacobs.
\newblock Predicting ground-level scene layout from aerial imagery.
\newblock In {\em IEEE Conference on Computer Vision and Pattern Recognition
  (CVPR)}, 2017.

\bibitem{han2017stackgan}
Han Zhang, Tao Xu, Hongsheng Li, Shaoting Zhang, Xiaogang Wang, Xiaolei Huang,
  and Dimitris Metaxas.
\newblock Stackgan: Text to photo-realistic image synthesis with stacked
  generative adversarial networks.
\newblock In {\em {ICCV}}, 2017.

\bibitem{CycleGAN2017}
Jun-Yan Zhu, Taesung Park, Phillip Isola, and Alexei~A Efros.
\newblock Unpaired image-to-image translation using cycle-consistent
  adversarial networkss.
\newblock In {\em Computer Vision (ICCV), 2017 IEEE International Conference
  on}, 2017.

\bibitem{zhu2018generative}
Xinge Zhu, Zhichao Yin, Jianping Shi, Hongsheng Li, and Dahua Lin.
\newblock Generative adversarial frontal view to bird view synthesis.
\newblock In {\em 2018 International Conference on 3D Vision (3DV)}, pages
  454--463. IEEE, 2018.

\end{thebibliography}
}

\clearpage





\title{Bridging the Domain Gap for Ground-to-Aerial Image Matching: \\
Supplementary Materials}

\author{Krishna Regmi
and
Mubarak Shah\\
Center for Research in Computer Vision, University of Central Florida\\
{\tt\small krishna.regmi7@gmail.com, shah@crcv.ucf.edu}
}

\setcounter{section}{0}
\renewcommand*{\theHsection}{chX.\the\value{section}}

\twocolumn[{%
\renewcommand \twocolumn[1][]{#1}%
\maketitle
\centering
\vspace{10pt}
\includegraphics[width=0.47\textwidth]{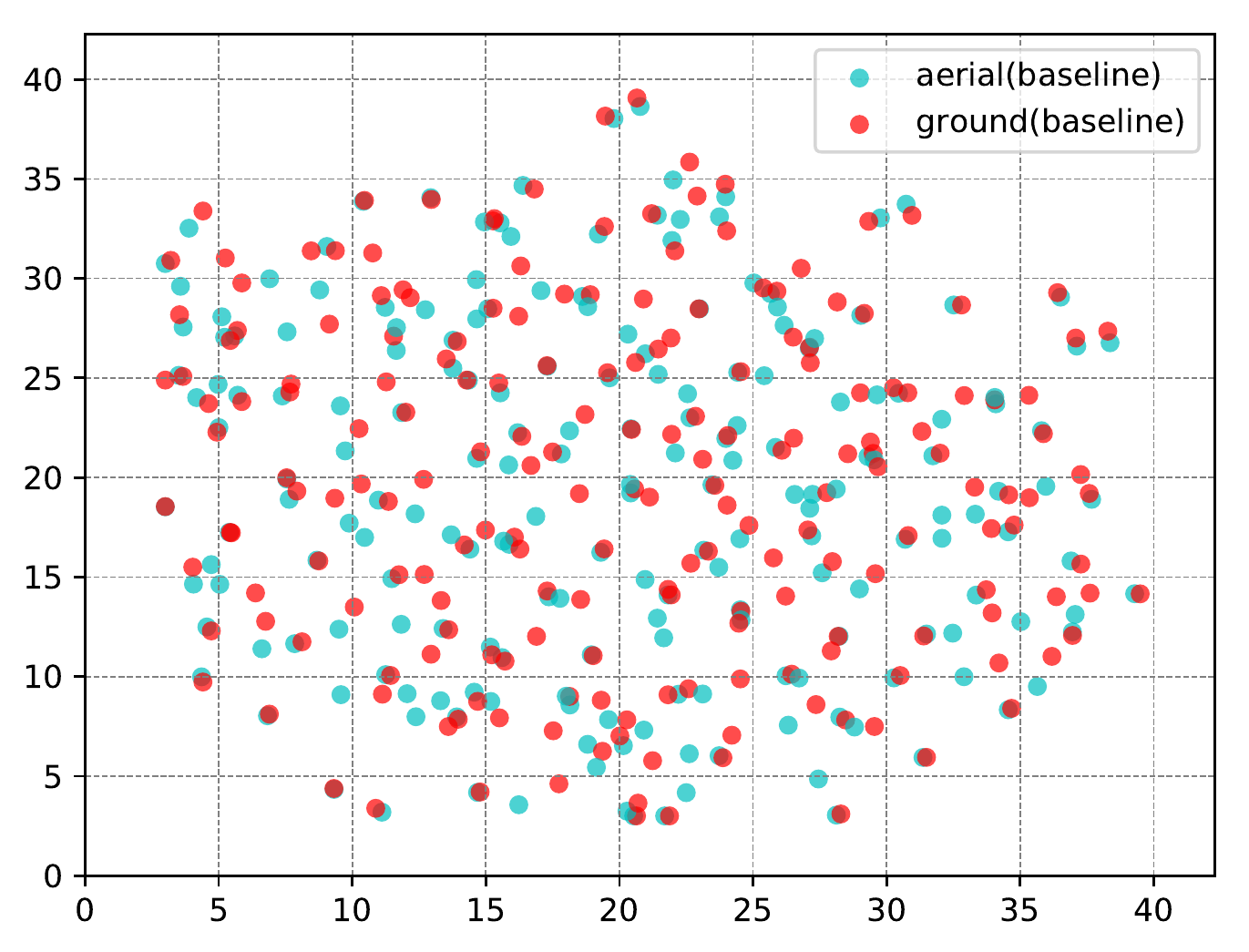} \hspace{5pt}
\includegraphics[width=0.47\textwidth]{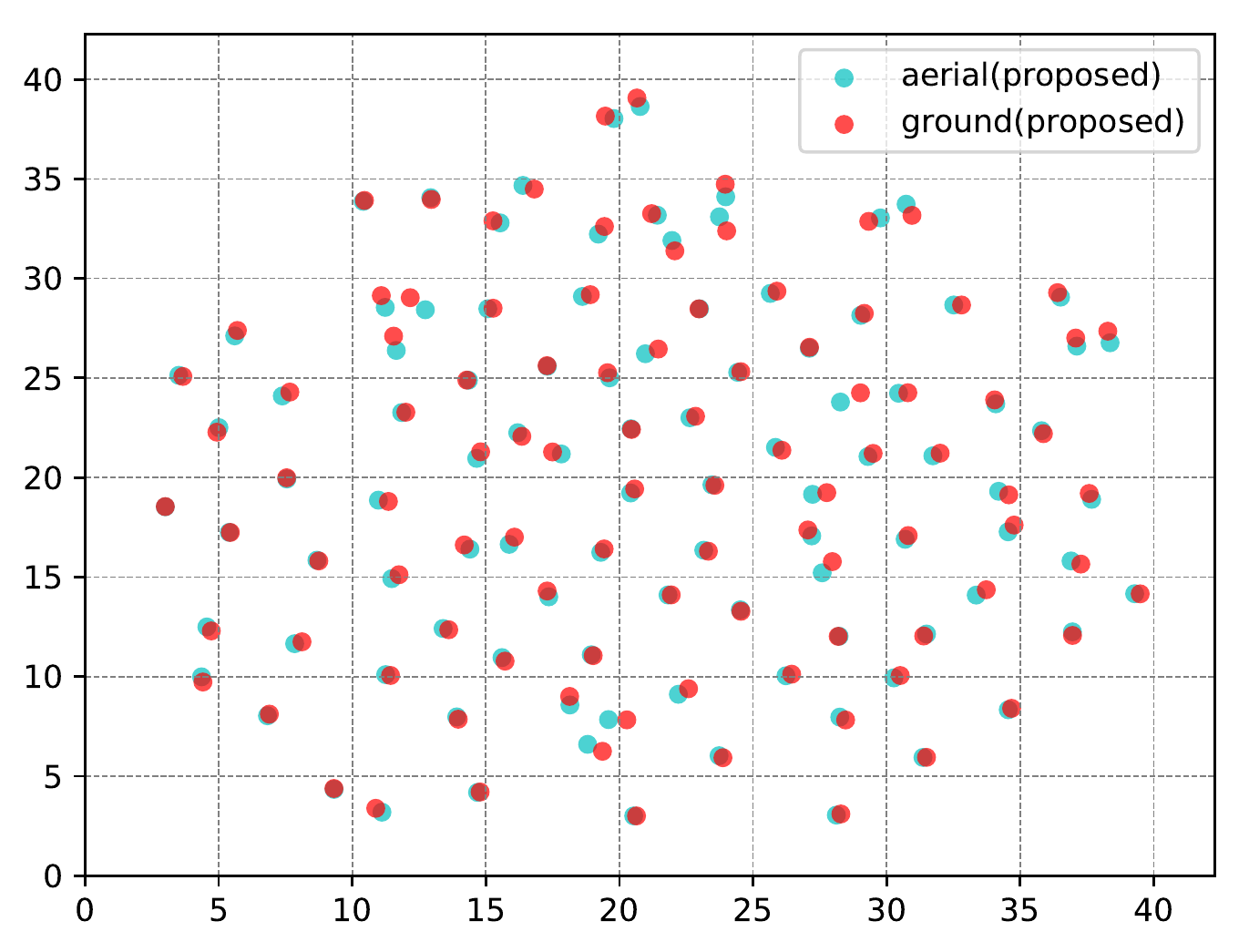}
\captionof {figure} {\small \label{fig:tsne}T-SNE \cite{maaten2008visualizing} visualization of aerial and ground image features obtained using the two-stream baseline (left) and the proposed feature fusion method (right) for 100 test images on the CVUSA dataset \cite{zhai2017crossview}.}
\vspace{20pt}
}]


This supplementary material provides an in-depth analysis of our proposed method for tackling the cross-view image matching on the CVUSA dataset \cite{zhai2017crossview}. We apply t-SNE \cite{maaten2008visualizing} to the feature representations obtained from two-stream baseline and the proposed feature fusion method to project them to  two-dimensional representations and visualize them. 
We provide some qualitative results for failure examples.   
We also explore the reverse problem of aerial-to-ground image matching by using the proposed joint feature learning and  feature fusion method. 
Furthermore, we present some qualitative results on the OP dataset.

\section{Visualization and Interpretation of Features}

In Figure \ref{fig:tsne}, we visualize the aerial and ground image features obtained using the two-stream baseline and the proposed feature fusion methods for 100 images on the CVUSA dataset \cite{zhai2017crossview}. 
The feature representation for each image is a 1000-dimensional vector and we apply t-SNE to learn their two-dimensional embeddings for ease of visualization. 
The red and cyan circles close to each other or with some overlap represent the features for the ground
image and its corresponding ground-truth aerial image respectively in the subplots.

The scatter-plot for features obtained using the two-stream baseline trained on ($I_g$, $I_a$) pairs is shown on the left. We observe that, for each image pair, there is less overlap between the aerial and the ground image features. We also notice that the features from different image pairs are located close to each other, with some instances of red circles overlapping each other.

The scatter-plot for the representations obtained using the proposed feature fusion method trained on image triads ($I_g$, $I_a$, $I_{a'}$) is shown on the right subplot. We observe higher overlap between the features for ground and corresponding aerial image pairs. At the same time, we observe greater separation between the feature embeddings for different image samples. 

Thus, the use of synthesized aerial images in our proposed Feature Fusion method are successful in bringing the feature representations of aerial images closer to the representations of ground images and bridging the domain gap between the images from these two drastically different views to improve the matching accuracy.



\begin{figure}[t]
\begin{center}
  \includegraphics[width=0.99\linewidth]{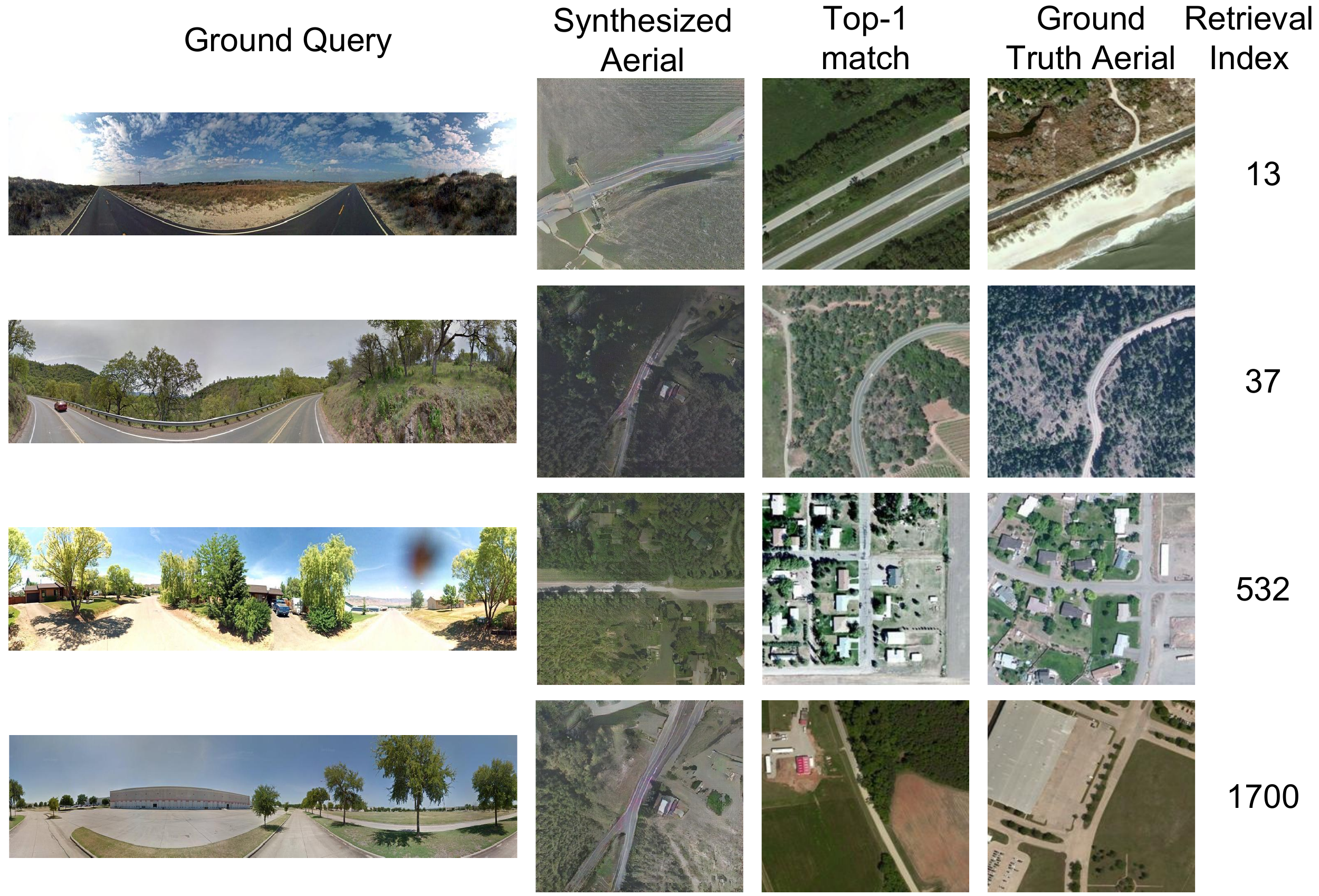}
\end{center}
  \caption{\small Some examples of failure cases. The numbers on the right show the position where the ground-truth aerial images were retrieved.}
\label{fig:failure_cases}
\end{figure}


\section{Failure Examples}
We present some failure cases for the proposed Feature Fusion method in Figure \ref{fig:failure_cases}. In each row, We respectively present the query ground image,  corresponding synthesized aerial image, image retrieved at Top-1 position, ground-truth aerial image and a number representing the position where the ground-truth aerial image was retrieved. 

Row 1 shows that ground truth aerial image consisting of water body in lower right section of the image. The ground image does not provide any information regarding water, so the image matching is challenging. The ground truth is retrieved at 13$^{th}$ position.

In Row 2, we can observe that the top match and ground-truth aerial images are very similar to each other. Also, the matched image has similar color distribution to query image than the ground-truth aerial image. The problem arises because the aerial and ground image pairs in the dataset are captured at different times, so have some visual differences.

Row 3 shows an example where the aerial image has houses which are not captured in street-view images due to occlusion by trees. The impact can also be observed in the corresponding synthesized image which doesn't contain houses. 

Row 4 shows that the street-view image contains a building at far distance. The building covers large region in ground-truth aerial image, which is difficult to comprehend from the street-view image. Also, this is a rare situation in the dataset with large building. So, the method fails badly, retrieving the ground-truth image at position 1700.


\section{Aerial-to-Ground Image Matching}
We conducted experiments for the reverse problem of Aerial-to-Ground image matching. Here, the aerial image is the query, and we attempt to find the matching ground panorama. First, we use GANs to synthesize ground level panoramas from the aerial images and then use the synthesized images in the proposed joint feature learning and feature fusion methods.


\begin{table}[t]
 \small
  \centering
  \renewcommand{\tabcolsep}{1.2mm}  
  \caption{\small Image matching performance in terms of Top-1, Top-10 and Top-1\% recall on CVUSA Dataset \cite{zhai2017crossview} for aerial-to-ground matching.}
  \vspace{-5pt}
  \label{tab:one_ten_1percent_rev}
    \begin{tabular}{l|ccc}
        \toprule
        \textbf{Method} & \textbf{Top-1} &\textbf{Top-10} & \textbf{Top-1\%} \\
        \midrule
        \midrule

        Two-stream baseline ($I_{g'}$, $I_g$) & $ 15.04\% $& $  37.31\% $  & $  67.99\% $\\

         Two-stream baseline ($I_a$, $I_g$) & 16.99\%  &$ 47.06\%  $ & 82.11\%   \\


        Joint Feat. Learning ($I_{g'}$, $I_g$) & 16.46\%  & 50.26\%  & 86.26\%    \\

        Joint Feat. Learning ($I_a$, $I_g$) & 27.39\%  & 65.29\%  & 91.46\%    \\

            
        Feature Fusion & {\bf 44.99\%}  & {\bf 79.37\%}  & \bf{95.66\% }   \\

        \bottomrule
    \end{tabular}
\end{table}

\begin{figure}[t]
\begin{center}
   \includegraphics[width=0.9\linewidth]{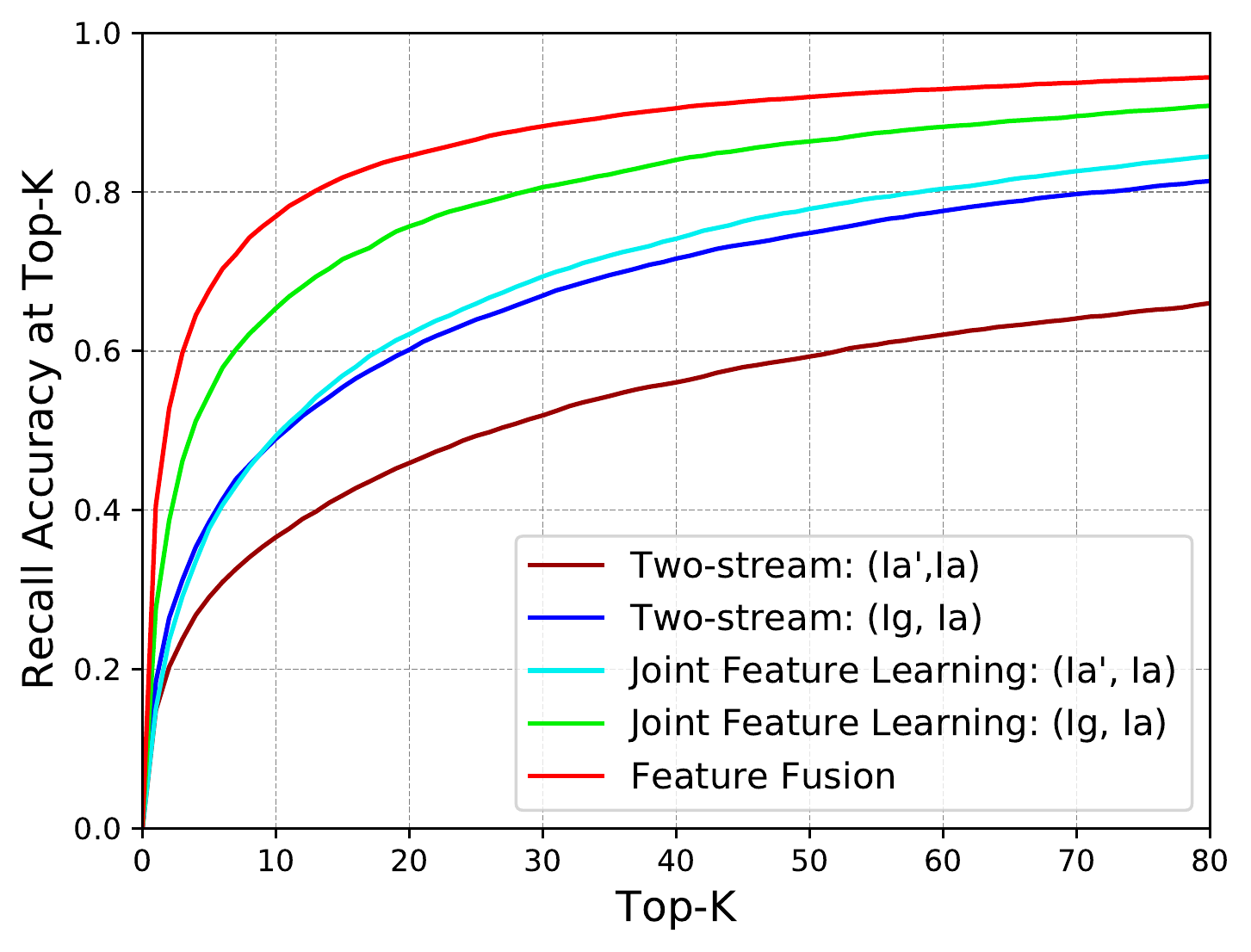}
\end{center}
   \caption{\small Comparison of our methods with the baselines on CVUSA dataset \cite{zhai2017crossview} for reverse problem of aerial-to-ground image matching.}
\label{fig:auc_reverse}
\end{figure}

\begin{figure*}[]
\begin{center}
  \includegraphics[width=0.99\linewidth]{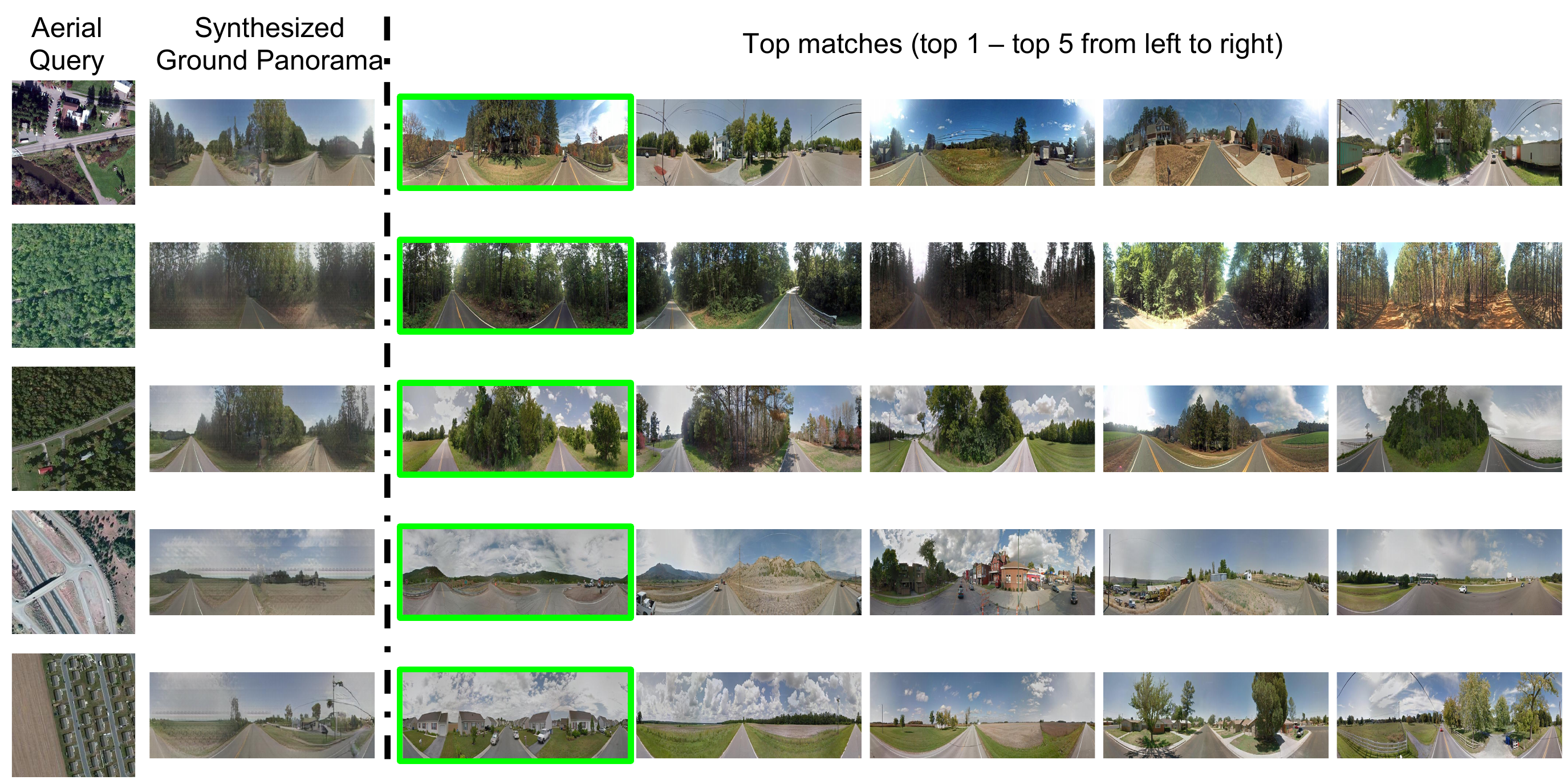}
\end{center}
  \caption{\small Qualitative Results on CVUSA dataset \cite{zhai2017crossview} for aerial-to-ground image matching. Images with green borders are the ground-truth panoramas for the corresponding query images.}
\label{fig:qual_a2g}
\end{figure*}


\begin{figure*}[]
\begin{center}
  \includegraphics[width=0.99\linewidth]{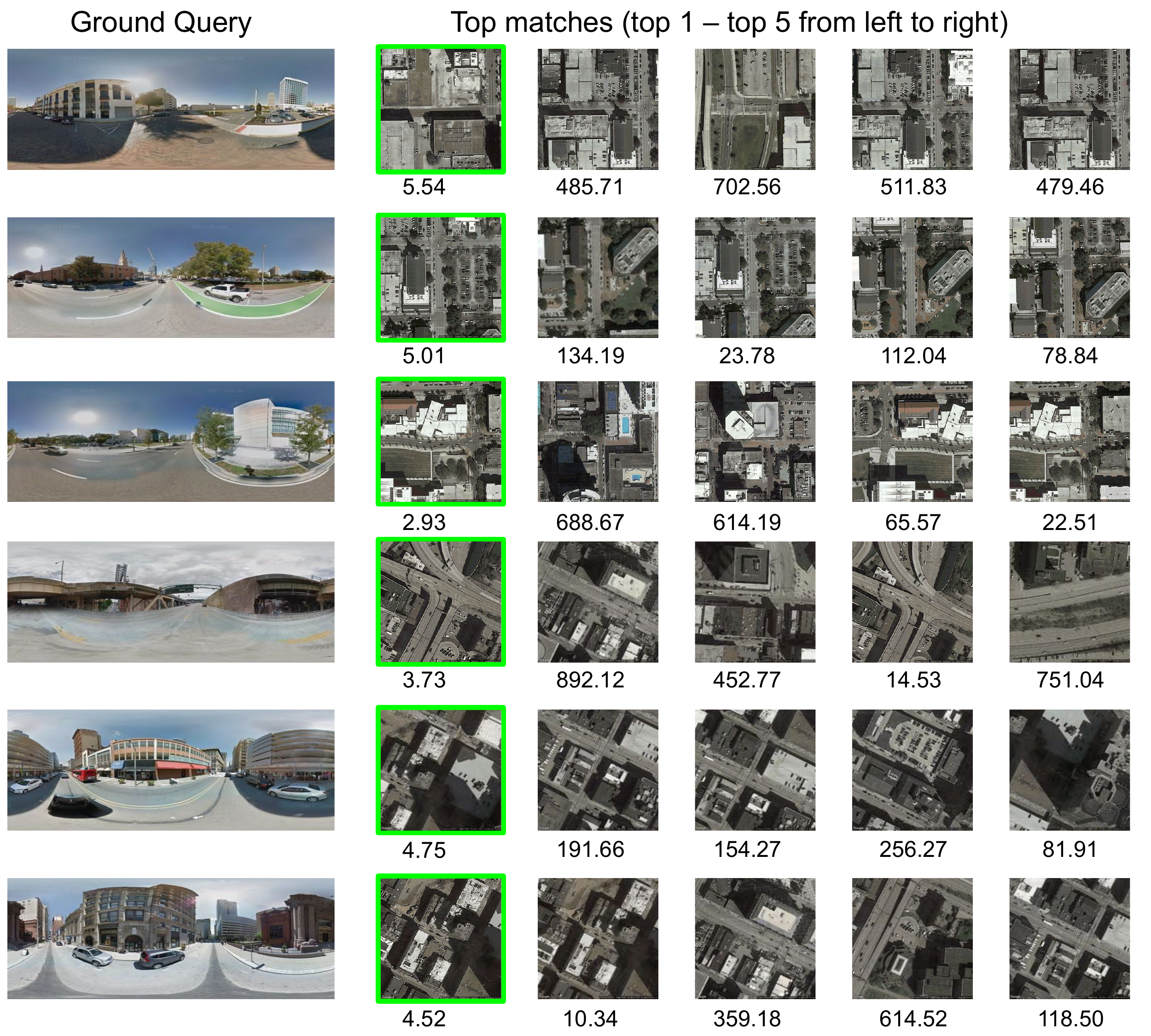}
\end{center}
  \caption{\small Cross-view image retrieval examples on the OP Dataset. Ground-truth aerial images are shown in green boxes. The number below each aerial image is its distance in meters from the query image. The first three rows present the images from Orlando and the next three rows of images are from Pittsburgh. 
}
\label{fig:sample_op}
\end{figure*}

We conduct experiments for two-stream baselines, joint feature learning and feature fusion methods. The top-1, top-10 and top-1\% accuracies are reported in Table \ref{tab:one_ten_1percent_rev}. We obtain results similar to the numbers reported in the main paper for ground-to-aerial image matching. 
We also plot the top-K recall for K = 1 to 80 for the proposed method compared to the baselines in Figure \ref{fig:auc_reverse}. This affirms that our method can be applied for image matching in both directions.

The qualitative results for aerial-to-ground image matching are shown in Figure \ref{fig:qual_a2g}. The query aerial image, synthesized ground panorama followed by the three closest matches are visualized. The ground-truth panorama are shown with the green borders. We can also observe that the synthesized ground panorama are successful in transforming the semantic information from aerial to ground domain.


\section{OP Dataset}
The existing public datasets on cross-view image matching do not provide geo-information. Also, the images on the CVUSA dataset are collected from the rural areas that largely cover land and vegetation as shown  
in Figures \ref{fig:failure_cases} and \ref{fig:qual_a2g}. To compensate those issues, we collect a new dataset of cross-view image pairs. The images cover urban areas of Orlando and Pittsburgh. Figure \ref{fig:sample_op} shows the example images of this dataset. We can observe that this dataset contains images of mainly urban areas with buildings and roads and less vegetation, contrasting to the CVUSA dataset. 

We conduct experiments on the OP dataset and provide more qualitative results in Figure \ref{fig:sample_op}. The number below each aerial image represents its distance in meters from the query ground image. We observe that though the aerial images look very similar to each other, the proposed feature fusion method is able to retrieve the ground-truth aerial image as the closest matching image. 
The quantitative evaluation is provided in the main paper. We obtain similar results for top-1 accuracies on both the CVUSA and the OP dataset. 
This consolidates the fact that the proposed method generalizes well on both rural and urban datasets.



\end{document}